\documentclass[10pt,journal,compsoc]{IEEEtran}
\usepackage{microtype}
\usepackage{graphicx}
\usepackage{booktabs} 

\usepackage{hyperref}

\usepackage{multirow}
\usepackage{amsmath}
\usepackage{amsthm}
\usepackage{amssymb}
\usepackage{bm}
\usepackage{algorithm}
\usepackage{algorithmic}
\usepackage{graphicx,subfig}
\usepackage{xcolor}
\usepackage{color}
\usepackage[table]{xcolor}  
\usepackage{colortbl}       

\newcommand{\norm}[1]{\left\lVert#1\right\rVert}

\def \R {\mathbb{R}}

\def \vv {\bm{v}}

\newtheorem{assumption}{Assumption}

\newtheorem{lemma}{Lemma}

\newtheorem{theorem}{Theorem}

\newtheorem{remark}{Remark}

\ifCLASSOPTIONcompsoc
  \usepackage[nocompress]{cite}
\else
  \usepackage{cite}
\fi
\ifCLASSINFOpdf
\else
\fi
\begin{document}
%
\title{Federated Learning via Variational Bayesian Inference: Personalization, Sparsity and Clustering}
%
%
%
%

\author{Xu Zhang,~\IEEEmembership{Member,~IEEE,}
        Wenpeng Li,
        Yunfeng Shao,~\IEEEmembership{Member,~IEEE,}
        Yonglin Liu,
        Kaiwen Zhou,
        Yinchuan Li,~\IEEEmembership{Member,~IEEE}
\IEEEcompsocitemizethanks{\IEEEcompsocthanksitem  Corresponding author: Yinchuan Li.\protect \\
\IEEEcompsocthanksitem X. Zhang is with the School of Artificial Intelligence, Xidian University, China and LSEC, Academy of Mathematics and Systems Science, Chinese Academy of Sciences, Beijing, China. Email: zhang.xu@xidian.edu.cn. \protect \\
\IEEEcompsocthanksitem W. Li, Y. Shao, Y. Liu, K. Zhou, and Y. Li are with Noah’s Ark Lab, Huawei, China. Emails:  wenpengli.ai@gmail.com,  shaoyunfeng@huawei.com, smarthehelyl@gmail.com, kwzhou@cse.cuhk.edu.hk, yinchuan.li.cn@gmail.com.}

}
\IEEEtitleabstractindextext{%
\begin{abstract}
Federated learning (FL) is a promising framework that models distributed machine learning while protecting the privacy of clients. However, FL suffers performance degradation from heterogeneous and limited data. To alleviate the degradation, we present a novel personalized Bayesian FL approach named pFedBayes. By using the trained global distribution from the server as the prior distribution of each client, each client adjusts its own distribution by minimizing the sum of the reconstruction error over its personalized data and the KL divergence with the downloaded global distribution. Then, we propose a sparse personalized Bayesian FL approach named sFedBayes to enhance the inference efficiency. To overcome the extreme heterogeneity in non-i.i.d. data, we propose a clustered Bayesian FL model named cFedbayes by learning different prior distributions for different clients. Theoretical analysis gives the generalization error bound of three approaches and shows that the generalization error rates of the proposed approaches achieve minimax optimality up to a logarithmic factor. Moreover, cFedBayes achieves a cluster-level generalization error bound, rather than a single uniform bound in pFedBayes. Numerous experiments demonstrate that the proposed approaches have better performance than other advanced personalized methods on private models in the presence of heterogeneous and limited data.
\end{abstract}

}

\maketitle

\IEEEdisplaynontitleabstractindextext

%
\IEEEpeerreviewmaketitle

\IEEEraisesectionheading{\section{Introduction}\label{sec:introduction}}

%
%
%
%

\IEEEPARstart{F}{ederated} Learning (FL) is a promising distributed machine learning model that can achieve data security and privacy protection\cite{mcmahan2017communication,Li2020FL}. Since each user's data does not leave its local client during model training, FL has been widely used in various applications such as medical diagnostics, autonomous driving, e-commerce, voice assistants and input prediction. The vanilla FL assumes that data from different clients are independent and identically distributed (i.i.d.) and are sufficient to train a well-performing model. However, data from practical applications are not perfect, but rather heterogeneous (non-i.i.d.) and limited: the heterogeneity of private data is due to the differences in users' habits, locations and preferences and the shortage of data is due to the high cost of data collection in practice.

To address the problems caused by heterogeneous data, researchers have proposed various Personalized Federated Learning (PFL) models \cite{li2018federated,t2020personalized,hanzely2020federated,t2020personalized,li2021personalized}, which exploit the correlation between the server's global parameters and the clients' local parameters to design customizations for each client. To further overcome the problems caused by extremely heterogeneous data, clustered federated learning (CFL) models are proposed \cite{ghosh2020efficient,xie2020multi,huang2021personalized}, where clients are assumed to belong to different clusters. Although these PFL and CFL models have gained performance improvements in heterogeneous data, these models perform poorly in the absence of sufficient data due to the overfitting. To mitigate the model overfitting of vanilla FL, a Bayesian Neural Network (BNN) is combined with FL to represent the network parameters of the server through probability distributions \cite{chen2020fedbe,thorgeirsson2021probabilistic}. However, these Bayesian Federated Learning (BFL) models perform poorly in the presence of heterogeneous datasets from different clients. A natural question is how to simultaneously address the challenges of heterogeneous and limited data.

In this paper, we propose a two-level optimization framework to incorporate BNN into PFL and CFL via variational Bayesian inference. Different from traditional BNNs, no prior distribution is assumed for each parameter on the end devices and the trained global distribution serves as the prior distribution. In this way, we avoid the shortcomings that the assumed prior distribution usually is not compatible with the true distribution. Furthermore, the introduction of BNNs enables our models to measure the output's uncertainty, which is useful in a range of robustness-critical applications such as visual security, medical diagnosis, internet of vehicles, and e-commerce \cite{jospin2020hands}.

\subsection{Main Contributions}

This paper presents three novel two-level personalized BFL algorithms via variational inference, i.e., pFedBayes, sFedBayes and cFedBayes. Among these algorithms, BNNs are used for clients' and server's  neural networks in these algorithms. In general, for the three proposed algorithms, the server aims to update the global distribution by aggregating the uploaded local distributions while the clients desire to minimize the sum of the reconstruction error over their own data and the KL divergence with the updated global distribution. 

We first propose a federated Bayesian learning method named pFedBayes to achieve personalization in the presence of limited data. All the parameters of the neural network are assumed to follow Gaussian distribution. Theoretical guarantee states that the average generalization error convergence rate achieves {\em minimax optimality}  up to a logarithmic term.

Then, \textcolor{black}{to accelerate the inference procedure, we develop a variant named sFedBayes that sparsifies the structure of BNNs by assuming all parameters follow a Bernoulli–Gaussian distribution. In this setting, pFedBayes can be viewed as a special case of sFedBayes when the sparsity level is one, and an appropriate sparsity achieves a favorable trade-off between inference efficiency and model performance. Theoretical analysis further shows that sFedBayes attains the {\em minimax optimal} convergence rate of the average generalization error in the sparse regime.}

Next, we present a clustered  Bayesian federated learning algorithm named cFedBayes to address the extreme statistical diversity among clients. Different from pFedBayes, which updates only one global distribution,  the server in cFedBayes updates several global distributions and each client chooses the best distribution that matches its private data from the global distributions in each round. Theoretical analysis indicates that cFedBayes provides 
cluster-level adaptive generalization guarantees, 
rather than a single overall bound across all clients as in pFedBayes.

Finally, we provide the corresponding stochastic gradient algorithms for the three proposed methods by updating the clients' distributions and the global distribution alternatively. Then we make the performance comparisons with other advanced approaches under heterogeneous datasets with different sizes. Extensive experiments demonstrate that the proposed methods perform better than current advanced algorithms in the presence of heterogeneous and limited data. In addition, simulations show that sparsification can improve test accuracy.

This paper is an extension of our previous conference
version \cite{zhang2022personalized}, and there are three major improvements over the preliminary one: 1. A sparse personalized Bayesian federated learning model named sFedBayes has been proposed to simplify the network structure \textcolor{black}{and enhance the inference efficiency}, and the corresponding generalization bound has been given; 2. A clustered Bayesian federated learning model named cFedBayes has been proposed to overcome the extreme statistical diversity among clients and \textcolor{black}{the generalization error bound is established at the cluster level, providing adaptive guarantees compared to the global averaged bound in pFedBayes}; 3. Extensive experiments have been added to demonstrate the superior performance of the proposed models when compared to current SOTA models.

\subsection{Related Works}
The proposed Bayesian federated learning  models are a hybrid of personalized/clustered federated learning and Bayesian neural networks, which is highly related to federated learning, personalized federated learning, and clustered federated learning.

\textit{Federated learning} was first proposed by the Google group in 2017, where Federated Averaging (FedAvg) was provided to ensure data security in distributed systems \cite{mcmahan2017communication}. Then Stich et al. \cite{stich2018local} proved the convergence of FedAvg for i.i.d. data. FedAvg started a new era for privacy-preserving machine learning. To meet diverse requirements from various circumstances, multiple alternative versions of FedAvg were provided. Researchers proposed many communication-efficient methods, including approximate Newton's algorithm \cite{li2019feddane}, primal-dual algorithm \cite{zhang2020fedpd} and one-shot averaging algorithm \cite{guha2019one}. To decrease the burden of storage and communication simultaneously, sparse methods \cite{sattler2019robust,rothchild2020fetchsgd} and quantized methods \cite{dai2019hyper,reisizadeh2020fedpaq} are integrated into federated learning. Since these methods aim to learn a global model for all clients, they have a subpar performance in the absence of i.i.d. data.

To cross the barrier posed by non-i.i.d. data, a great many \textit{personalized federated learning} are put forth, which include personal tailor (PT) methods \cite{li2018federated,arivazhagan2019federated, hanzely2020federated,t2020personalized,li2021personalized}, 
federated multi-task learning (FMTL) methods \cite{smith2017federated,sattler2021clustered} and federated meta-learning (FML) method \cite{chen2018federated,fallah2020personalized}. PT methods create a personal model for each client. As an illustration, Li et al. introduced a proximal term to the objective function and proposed an algorithm called Fedprox \cite{li2018federated}; Huang et al. put forth FedAMP and HeurFedAMP by creating an attentive message passing system to encourage cooperation among comparable clients\cite{huang2021personalized}; Arivazhagan et al. combined the cloud model and personalized layer to capture the personalization for each client \cite{arivazhagan2019federated}; Hanzely and Richt\'arik  proposed a mixture model by jointly minimizing the global loss and local loss \cite{hanzely2020federated}.
As a special type of PT method, the bilevel modeling approach achieves personalization by solving a bilevel optimization problem, which is made of the global subproblem and local subproblems. For instance, pFedMe used the Moreau envelope as a regularization term in the client-level subproblem to decouple the clients’ parameter from the server’s parameter \cite{t2020personalized}.
Smith incorporated multi-task learning into FL and put forth a corresponding algorithm \cite{smith2017federated}. In terms of the FML methods, Fallah et al. proposed the per-FedAvg approach by jointly creating a common model and then customizing it for all clients \cite{fallah2020personalized}. Despite their improvement on non-i.i.d. data, the above algorithms might overfit for insufficient data.

To further overcome the challenge of data heterogeneity, researchers proposed several \textit{clustered federated learning} approaches\cite{ghosh2020efficient,xie2020multi,sattler2021clustered,xing2022big,ma2022convergence}, where the clients are assumed to be in different clusters. Ghosh et al. proposed an iterative federated clustering algorithm (IFCA) by alternating between identifying the cluster membership of each client and optimizing the cluster models \cite{ghosh2020efficient}. Xie et al. proposed a multi-center FL algorithm named federated SEM (FeSEM) by solving a joint optimization and assigning the center for each client \cite{xie2020multi}.
Sattler et al. proposed a federated multitask model named FedCFL by using the geometric properties of the loss function to separate the clients \cite{sattler2021clustered}.
Xing et al. provided a bilevel optimization enhanced graph-aided federated learning (BiG-Fed) approach, where the graph structure is incorporated to collaboratively train heterogeneous models \cite{xing2022big}. 
Like PFL algorithms, CFL algorithms also cannot address the overfitting problem caused by limited data.

To suppress the overfitting caused by insufficient data, several \textit{federated Bayesian learning} algorithms were proposed.
The Bayesian nonparametric federated learning (BNFed) using neural parameter matching was suggested \cite{yurochkin2019bayesian}.
A new aggregation approach via Bayesian ensemble on the server's side was introduced \cite{chen2020fedbe}. A variant of FedAvg was proposed by incorporating Gaussian distribution to each parameter and aggregating the mean and variance of the local models \cite{thorgeirsson2021probabilistic}. The models above, however, perform poorly for heterogeneous data. FedPA is an approximate posterior inference approach that infers the server's posterior via the average of the clients' posteriors \cite{al2021federated}, but it is not suitable for PFL. Achituve et al. proposed a PFL called pFedGP that jointly learns a common kernel function via deep learning and customizes the Gaussian process classifier using the private data from each client \cite{achituve2021personalized}. Due to the fact that the same kernel function is used by all clients, performance may suffer when data volatility is considerable. A BFL model called FOLA uses the Gaussian approximate posterior distribution of the server as the local prior distribution. Compared with our methods, our approaches are based on a Bayesian bilevel optimization while FOLA is based on maximum a posteriori estimation. Moreover, whereas our approaches provide theoretical analysis, FOLA is short of theoretical analysis.

\begin{figure}[!t]
      \centering
      \centerline{\includegraphics[width=1.1\linewidth]{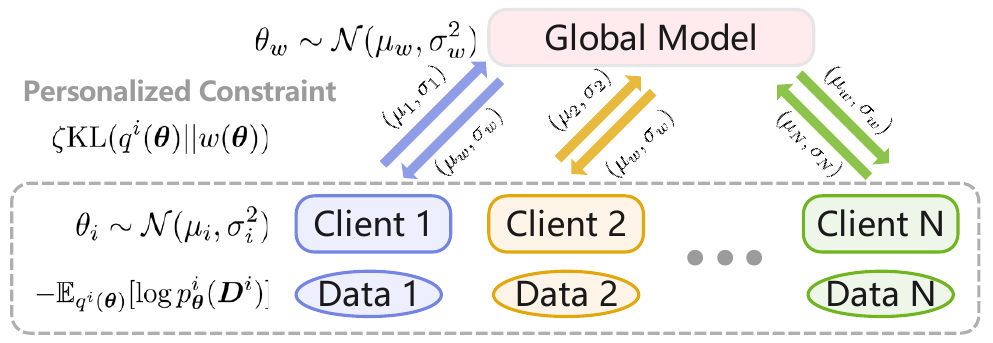}}
  
\caption{System diagram of the personalized Bayesian federated learning model under Gaussian assumptions.}
    \label{fig:Systemdiagram}
\end{figure}

\section{Problem Formulation}

In this section, we provide the problem formulation for Bayesian Federated Learning. Consider a distributed system in Fig. \ref{fig:Systemdiagram}. There are $N$ clients and one server, where each client owns its personalized BNN. In each round, each client updates its personal distribution based on its data and the global distribution received from the server and then sends them to the server. The server aggregates the received local distribution parameters and then sends them to the clients. The $i$-th client has a dataset $\bm{D}^i=(\bm{D}_1^i,\ldots,\bm{D}_n^i)$ with $\bm{D}_j^i=(\bm{x}_j^i,\bm{y}_j^i), i=1,\ldots, N$. The data follow the model
\begin{equation}
    \bm{y}_j^i=f^i(\bm{x}_j^i)+\varepsilon_j^i,\, j=1,\ldots,n,\, \varepsilon_j^i \sim \mathcal{N}(0,\sigma_\varepsilon^2),
\end{equation}
where $\bm{x}_j^i\in \R^{r_0}$, $\bm{y}_j^i \in \R^{r_{L+1}}$ for $j=1,\ldots,n$, $i=1,\ldots,N$, $f^i(\cdot): \R^{r_0} \to \R^{r_{L+1}}$ represents an unknown nonlinear function to be approximated, $n$ represents the number of samples, and $\sigma_\varepsilon$ represents the variance of noise. Our purpose is to propose a PFL model to approximate the unknown functions $\{f_i\}_{i=1}^N$ in the presence of heterogeneous and limited data.

Motivated by the universal approximation theorem \cite{cybenko1989approximation}, we use the fully-connected deep neural network (DNN) to approximate the unknown functions $\{f_i\}_{i=1}^N$ and represent the DNN model for the $i$-th client with $f_{\bm{\theta}}^i$. For the ease of aggregation of the server, we assume that all clients and the server share the same DNN structure but have different network parameters. Let the common neural network have $L$ hidden layers and the number of neurons for each layer be $r_j, j=1,\ldots, L$, respectively. The DNN model is presented as
\begin{equation}
    f_{\bm \theta}(\bm{x})= \bm{W}_{L+1}\sigma(\bm{W}_L\sigma(\ldots \sigma(\bm{W}_1 \bm{x}+ \bm{b}_1) )+\bm{b}_{L})+\bm{b}_{L+1},
\end{equation}
where $\sigma(\cdot)$ represents the activation function, $\bm{x}\in \R^{r_0}$, $\bm{W}_j\in \R^{r_{j-1}\times r_j}$ and $\bm{b}_j \in \R^{r_j}$. In this case, $\bm{\theta} \in \R^T$ represents the  vector that stacks all weights $\{\bm{W}_j\}_{j=1}^{L+1}$ and biases $\{\bm{b}_j\}_{j=1}^{L+1}$. The length of $\bm{\theta}$ is denoted by $T=|\bm{\theta}|$. Define $\bm{r}=\{r_1,\ldots,r_L\}$. Suppose that all the entries in $\bm{\theta}$ are bounded, i.e., $\norm{\bm{\theta}}_\infty \le B$, where $B>0$ stands for an absolute constant.

Next, we establish a bilevel optimization problem to estimate the distribution of the network parameters $\{\bm \theta_{i}\}_{i=1}^N$. To have a better understanding of the proposed model, we review the standard Bayesian variational inference in a single system \cite{VIJordan1999,Blei2017ReviewVB}. Let $\pi (\bm{\theta}|\bm{D})$ be the posterior distribution over data $\bm{D}$ and $q( \bm{\theta}) \in \mathcal{Q}$ be the probability distribution to be learned. Variational inference (VI) seeks to learn the optimal distribution $q( \bm{\theta})$ that is closest to the posterior distribution $\pi (\bm{\theta}|\bm{D})$ in terms of KL divergence
\begin{equation} \label{pb: VI}
 \min_{q( \bm{\theta}) \in \mathcal{Q}} \mbox{KL}\big(q( \bm{\theta})||\pi (\bm{\theta}|\bm{D})\big).
\end{equation}

By applying Bayes theorem $\pi (\bm{\theta}|\bm{D}) \propto \pi (\bm{\theta}) p_{\bm{\theta}} (\bm{D})$, the optimization problem \eqref{pb: VI} is reformulated to
\begin{equation} \label{pb: VI_2}
 \min_{q( \bm{\theta}) \in \mathcal{Q}} - \mathbb{E}_{q(\bm{\theta})}[\log p_{\bm{\theta}}(\bm{D})] +\mbox{KL}(q(\bm{\theta})||\pi (\bm{\theta})),
\end{equation}
where $\pi (\bm{\theta})$ stands for the prior distribution and $p_{\bm{\theta}} (\bm{D})$ stands for the likelihood. Here, the former term $- \mathbb{E}_{q(\bm{\theta})}[\log p_{\bm{\theta}}(\bm{D})]$ represents the reconstruction error over the data $\bm{D}$ while the latter term serves as a regularization term with the prior distribution. One disadvantage of vanilla VI is that it is difficult to characterize a good prior distribution of the dataset, but the performance of VI is dependent on a good choice of the prior distribution.

\begin{figure}[!t]
      \centering
        \centerline{\includegraphics[width=0.9\linewidth]{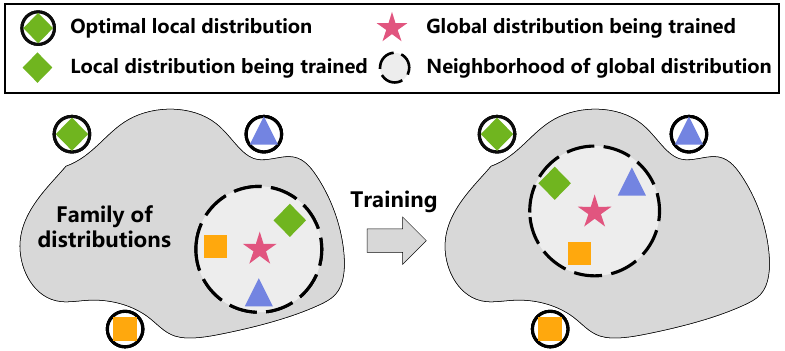}}
\caption{Distribution training of the personalized Bayesian federated learning model.}
    \label{fig:pFedBayes}
\end{figure}

To avoid the disadvantage in the centralized setting, we use the global distribution from the server as a surrogate of the prior distribution in the distributed setting. Let $w(\bm \theta) \in \mathcal{Q}_w$ be the probability distribution of the server's network parameters and $q^i(\bm \theta) \in \mathcal{Q}^i$ be the probability distribution of the $i$-th client's network parameters, where $\mathcal{Q}_w$ and $\mathcal{Q}^i$ represent the family of distributions for the server and the $i$-th client, respectively. Based on the vanilla VI \eqref{pb: VI_2}, we propose a new personalized Bayesian federated learning model by solving the following bilevel optimization problem
\begin{align}
\label{Master-1}
    \text {Server:}&~ \min _{w( \bm{\theta}) \in \mathcal{Q}_w}\left\{F(w)\triangleq \frac{1}{N} \sum_{i=1}^{N} F_{i}(w)\right\},
\end{align}
\begin{align}
\label{Client-1}
   \text {Clients:}~ F_i(w)\triangleq \min _{q^i(\bm{\theta}) \in \mathcal{Q}^i}\Big\{&- \mathbb{E}_{q^i(\bm{\theta})}[\log p^i_{\bm{\theta}}(\bm{D}^i)] 
   \nonumber\\
   & + \zeta\, \mbox{KL}(q^i(\bm{\theta})||w(\bm{\theta})) \Big\}.
\end{align}
where $p^i_{\bm{\theta}}(\bm{D}^i)$ represents the likelihood for the $i$-th client and $\zeta \ge 1$ is a constant that balances the personalization and global aggregation. The distribution training of the proposed model is shown in Fig. \ref{fig:pFedBayes}, which presents the change of the local and global distributions in probability space after training.  The server aims to minimize its KL divergence with all the clients' distributions, whereas each client seeks to strike a balance between minimizing the reconstruction error and minimizing the KL divergence with the global distribution. 

\section{Personalized Bayesian Federated Learning with Gaussian Distribution}

In this section, we study the theory and algorithm of personalized Bayesian federated learning under Gaussian distribution. All elements of the network parameters are assumed to be independent and identically distributed (i.i.d.) Gaussian variables, which is a common assumption in the literature \cite{Blundell2015,chen2020fedbe,thorgeirsson2021probabilistic}. Specifically, we suppose that each entry of the server's parameter vector $\bm \theta_{w}$ satisfies
\begin{equation}\label{wstar1}
\theta_{w,m} \sim \mathcal{N}(\mu_{w,m}, \sigma^2_{w,m}) ,~m=1,\ldots, T,
\end{equation}
where $\mu_{w,m}$ represents the mean and $\sigma_{w,m}$ represents the standard deviation of the $m$-th entry of the server. And suppose that each entry of the $i$-th client's parameter vector $\bm \theta_{i}$ follows
\begin{equation}\label{ptheta1}
\theta_{i,m}  \sim  \mathcal{N}(\mu_{i,m}, \sigma^2_{i,m})  , ~m=1,\ldots, T, 
\end{equation}
where $\mu_{i,m}$ represents the mean and $\sigma_{i,m}$ represents the standard deviation of the $m$-th entry of the $i$-th client.

Based on the above assumptions, we can give a closed-form result for the KL divergences of $q^i(\bm{\theta})$ and $w(\bm{\theta})$, which is very useful in the following analysis and algorithm
\begin{align} \label{kl_q_w}
    &~\mbox{KL}(q^i(\bm{\theta})||w(\bm{\theta})) \nonumber\\
    =&~\mbox{KL}\left(\prod^{T}_{m=1}\mathcal{N}(\mu_{i,m}, \sigma^2_{i,m}) \Bigl|\Bigr| \prod^T_{m=1}\mathcal{N}(\mu_{w,m}, \sigma^2_{w,m})\right) \nonumber\\
    =&~\sum_{m=1}^T  \mbox{KL}(\mathcal{N}(\mu_{i,m}, \sigma^2_{i,m})||\mathcal{N}(\mu_{w,m}, \sigma^2_{w,m}))  \\
    =&~ \frac{1}{2} \sum_{m=1}^{T} 
    \left[ \log \left( \frac{\sigma_{w,m}^2}{\sigma_{i,m}^2} \right) + \frac{\sigma_{i,m}^2+ (\mu_{i,m}-\mu_{w,m})^2}{\sigma_{w,m}^2} -1
    \right].\nonumber
\end{align}

\subsection{Theoretical Analysis}

This subsection gives the theoretical guarantee for pFedBayes by analyzing the average generalization error bound, which presents that pFedBayes achieves the minimax error rate. The related proofs are delayed in the Appendices. 

First of all, we provide some useful assumptions. The widths of all layers in BNNs are set to be the same \cite{polson2018posterior,bai2020efficient} and the activation function of each layer is set to be  $1$-Lipschitz continuous. 

\begin{assumption} \label{assump: equal-width}
The widths of all layers in BNNs are equal, i.e.,  $r_i=M,~i=1,\ldots,L$.
\end{assumption}

\begin{assumption} \label{assump: non-linear}
The  activation function in BNNs is $1$-Lipschitz continuous, i.e., $\norm{\sigma(x)-\sigma(y)}_2 \le \norm{x-y}_2$.
\end{assumption}

Since each element in $\{\bm{\theta_{i}}\}_{i=1}^N$ is upper bounded by $B$, its variance is upper bounded by $B^2$, which is used in the proof of Lemma \ref{lm: upper bound 2}.

\begin{assumption} \label{assump: sigma_bound}
The parameters $r_0,n,M,L$ are chosen suitably large such that 
\begin{equation} \label{eq: sigma_n}
    \sigma^2_{n}=  \frac{T}{8n} A \le B^2,
\end{equation}
where $H=BM$ and 
\begin{align} \label{eq: A}
&A=\log^{-1}(3r_0 M)\cdot(2H)^{-2(L+1)} \nonumber\\ &\left[\left(r_0 + 1 + \frac{1}{H-1}\right)^2 + \frac{1}{(2H)^2-1}
	+\frac{2}{(2H-1)^2}\right]^{-1}.
\end{align}
\end{assumption}

Next, we define the Hellinger distance
\begin{equation}
    d^2(P_{\bm{\theta}}^i, P^i) = \mathbb{E}_X\Bigl( 1 - \exp \{-[f_{\bm{\theta}}^i(X) - f^i(X)]^2/(8\sigma^2_{\epsilon}) \}\Bigr)
\end{equation}
and 
\begin{align}
\chi_n&= ((L+1)T/n)\log M + (T/n)\log(r_0\sqrt{n/T}), \label{chi}\\
\xi^i_n&= \inf_{\bm{\theta} \in {\Theta}(L, \bm{r}),\|\bm{\theta}\|_\infty\leq B}||f_{\bm{\theta}}^i-f^i||^2_\infty, \label{xi}\\
\varepsilon_n &= n^{-\frac{1}{2}}\sqrt{(L+1)T\log M + T\log(r_0\sqrt{n/T})}\log^\delta(n), \label{vare}
\end{align}
where $\delta>1$.

In order to give an upper bound of the average generalization error, we incorporate a constant in the bilevel optimization problem and the problem of the clients is reformulated as
\begin{align}
\label{Client-2}
   \text {Clients:}~ F_i(w)\triangleq \min _{q^i(\bm{\theta}) \in \mathcal{Q}^i} & \Big\{  \int_{\Theta} l_n(P^{i},P^{i}_{\bm{\theta}}) q^i(\bm{\theta}) d\bm{\theta} \nonumber\\
   &+ \zeta\,\mbox{KL}(q^i(\bm{\theta})||w(\bm{\theta})) \Big\},
\end{align}
where $l_n(P^{i},P^{i}_{\bm{\theta}})$ is the log-likelihood ratio of $P^{i}$ and $P^{i}_{\bm{\theta}}$
\begin{equation} \label{def: l_n}
    l_n(P^{i},P^{i}_{\bm{\theta}})=\log \frac{p^i(\bm{D}^i)}{p^{i}_{\bm \theta}(\bm{D}^i)}.
\end{equation}

Define $w^\star(\bm{\theta})$ as the solution of the server's problem \eqref{Master-1} and $\hat{q}^i(\bm{\theta})$ as the  solution for the $i$-th client's problem
\begin{align}
\label{prb: Master-pFedBayes}
     w^\star(\bm{\theta})=\arg \min _{w( \bm{\theta}) \in \mathcal{Q}_w}\left\{F(w)\triangleq \frac{1}{N} \sum_{i=1}^{N} F_{i}(w)\right\},
\end{align}
\begin{align} \label{prb: subproblem_optimal-pFedBayes}
    \hat{q}^i(\bm{\theta})
    =\arg \min_{q^i(\bm{\theta}) \in \mathcal{Q}^i} \Big\{ &\int_{\Theta} l_n(P^{i},P^{i}_{\bm{\theta}}) q^i(\bm{\theta}) d\bm{\theta} \nonumber\\
    &+ \zeta\, \mbox{KL}(q^i(\bm{\theta})||w^\star(\bm{\theta}))\Big\}.
\end{align}
Next, we will provide an upper bound for the average generalization error
\begin{align} \label{dist-pFedBayes}
  \frac{1}{N} \sum_{i=1}^{N} \int_{\Theta}   d^2(P_{\bm{\theta}}^i, P^{i}) \hat{q}^i(\bm{\theta}) d\bm{\theta}. 
\end{align}
with the defined parameters $\chi_n$, $\xi^i_n$ and $\varepsilon_n$. The analysis is divided into two steps: we first give the bound with the average of \eqref{prb: subproblem_optimal-pFedBayes} in Lemma \ref{lm: upper bound 1} and then provide the upper bound of the average of \eqref{prb: subproblem_optimal-pFedBayes} by using the optimality of $\hat{q}^i(\bm{\theta})$ in Lemma \ref{lm: upper bound 2}.

\begin{lemma} \label{lm: upper bound 1}
If Assumptions \ref{assump: equal-width} and \ref{assump: non-linear} hold, then the following inequality is satisfied with dominating probability
\begin{align} \label{neq: local_upperbound_sum}
    \frac{1}{N} & \sum_{i=1}^{N}  \int_{\Theta}  d^2(P_{\bm{\theta}}^i, P^i)  \hat{q}^i(\bm{\theta}) d\bm{\theta} \le \nonumber\\
    &\frac{1}{n} \Bigg\{\frac{1}{N} \sum_{i=1}^{N} \Bigg[ \frac{1}{\zeta} \int_{\Theta} l_n(P^{i}, P^{i}_{\bm{\theta}})\hat{q}^i(\bm{\theta}) d\bm{\theta}  \nonumber\\
    & \qquad \qquad \quad + \mbox{\rm KL}(\hat{q}^i(\bm{\theta})||w^\star(\bm{\theta})) \Bigg] \Bigg\} + C\varepsilon^{2}_{n}, 
\end{align}
where $\zeta \ge 1$ is a balance parameter and $C>0$ is a constant.
\end{lemma}

\begin{lemma} \label{lm: upper bound 2}
If Assumptions \ref{assump: equal-width}-\ref{assump: sigma_bound} hold, then the following inequality is satisfied with dominating probability
\begin{multline}
    \frac{1}{N} \sum_{i=1}^{N}  \left[  \int_{\Theta} l_n(P^{i},P^{i}_{\bm{\theta}}) \hat{q}^i(\bm{\theta}) d\bm{\theta} +  \zeta\, \mbox{\rm KL}(\hat{q}^i(\bm{\theta})||w^\star(\bm{\theta}))\right] 
    \\ \le n\left(  C' \zeta \chi_n +\frac{C''}{N} \sum_{i=1}^{N}\xi^i_n \right),
 \end{multline}
 where $\zeta \ge 1$ is a balance parameter and $C', C''$ are arbitrary diverging sequences.
\end{lemma}

By combining Lemmas \ref{lm: upper bound 1} and \ref{lm: upper bound 2}, we obtain the generalization error bound as follows.

\begin{theorem} \label{Thm_main-pFedBayes}
If Assumptions \ref{assump: equal-width}-\ref{assump: sigma_bound} hold, then the following generalization error upper bound  is satisfied with dominating probability
\begin{multline}
\label{neq: local_upperbound-pFedBayes}
  \frac{1}{N} \sum_{i=1}^{N} \int_{\Theta}  d^2(P_{\bm{\theta}}^i, P^i) \hat{q}^i(\bm{\theta}) d\bm{\theta}
    \\ \le  C \varepsilon^{2}_{n}
  + C'  \chi_n +\frac{C''}{N\zeta} \sum_{i=1}^{N}\xi^i_n,
\end{multline} 
where $\zeta \ge 1$ is a balance parameter, $C>0$ is a constant and $C', C''$ are arbitrary diverging sequences. 
\end{theorem}

\begin{remark} From \eqref{neq: local_upperbound-pFedBayes}, we see that the upper bound and the approximation error decrease as $\zeta$ increases. However, the degree of  customisation decreases as $\zeta$ increases, which is not what we want. Therefore, we should choose a suitable parameter $\zeta$ such that the degree of customisation and the average generalization error bound can be balanced.   
\end{remark}

As for Theorem \ref{Thm_main-pFedBayes}, the upper bound is divided into two kinds of error: the sum of the first two terms is the estimation error and the third term is the approximation error. According to the definitions of $\chi_n$ and $\varepsilon_{n}$, we obtain that the error bound gets smaller as the sample size $n$ increases and gets bigger as $T$ increases. However, the approximation error decreases with the increase of $T$ (or the width and depth). In order to balance the upper bound, we should choose a suitable $T$ as a function of the number of samples $n$.

Next, we discuss how to choose $T$ for $\beta$-Hölder-smooth functions $\{f^i\}$. Let the intrinsic dimension of data be 
$d$. From \cite[Corollary 6]{nakada2020}, the approximation error $\{\xi^i_n\}_{i=1}^N$ is upper bounded 
\begin{equation}
    \xi^i_n \le C_0 T^{-2\beta/d}, i=1,\ldots,N,
\end{equation}
where $C_0>0$ represents a constant related to $r_0$, $\beta$ and $d$.
By choosing $T=C_1 n^{d/(2\beta+d)}$ in Theorem \ref{Thm_main-pFedBayes}, the generalization bound is given by
\begin{align} \label{eq: convegencerate}
     \frac{1}{N} \sum_{i=1}^{N} \int_{\Theta}  d^2(P_{\bm{\theta}}^i, P^i) \hat{q}^i(\bm{\theta}) d\bm{\theta} \le C_2 n^{-\frac{2\beta}{2\beta+d}} \log^{2\delta'} (n),
\end{align}
where $C_1, C_2>0$ are constants related to $r_0$, $\beta$, $d$, $L$, $M$, $\zeta$ and $n$ and $\delta'>\delta>1$.

In the end, we demonstrate that the proposed pFedBayes achieves minimax optimality of the generalization error rate. Like \cite[Theorem 1.1]{bai2020efficient}, we can restate the upper bound using $L^2$ norm. Since the function $[1-\exp(-x^2)]/{x^2}$ is monotonically decreasing for $x>0$, for bounded functions $\norm{f^i}_\infty \le F$, $i=1,\ldots,N$, the following inequality holds
\begin{align}
    \frac{d^2(P_{\bm{\theta}}^i, P^i)}{\norm{f_{\bm{\theta}}^i(X) - f^i(X)}_{L^2}^2} \ge \frac{1-\exp(-\frac{4F^2}{8\sigma_\epsilon^2})}{4F^2} \triangleq C_F.
\end{align}
Combining with \eqref{eq: convegencerate}, we provide the upper bound of the generalization error using $L^2$ norm
\begin{multline}
\label{neq: local_upperbound_L_2}
  \frac{C_F}{N} \sum_{i=1}^{N} \int_{\Theta} \norm{f_{\bm{\theta}}^i(X^i) - f^i(X^i)}_{L^2}^2 \hat{q}^i(\bm{\theta}) d\bm{\theta} \\
  \le \frac{1}{N} \sum_{i=1}^{N} \int_{\Theta}  d^2(P_{\bm{\theta}}^i, P^i) \hat{q}^i(\bm{\theta}) d\bm{\theta} 
    \\ \le  C_2 n^{-\frac{2\beta}{2\beta+d}} \log^{2\delta'} (n).
\end{multline} 

Together with the minimax lower bound using $L^2$ norm in \cite[Theorem 8]{nakada2020}, we have
\begin{multline}
\label{neq: local_lowerbound_L}
 \inf_{\{\norm{f_{\bm{\theta}}^i}\le F\}_{i=1}^{N}} \sup_{\{\norm{f^i}_\infty\le F\}_{i=1}^N} \frac{C_F}{N} \sum_{i=1}^{N} \\ \int_{\Theta} \norm{f_{\bm{\theta}}^i(X^i) - f^i(X^i)}_{L^2}^2  \hat{q}^i(\bm{\theta}) d\bm{\theta}  \ge C_3 n^{-\frac{2\beta}{2\beta+d}},
\end{multline} 
where $C_3>0$ is a constant.

Since the minimax lower bound \eqref{neq: local_lowerbound_L} matches the upper bound \eqref{neq: local_upperbound_L_2}, we come to a conclusion that the convergence rate of the average generalization error of pFedBayes is minimax optimal up to a logarithmic term for bounded functions $\{f^i\}_{i=1}^{N}$ and $\{f^i_{\bm{\theta}}\}_{i=1}^{N}$.

\begin{algorithm}[!t] \small

\begin{minipage}{\linewidth}
	\caption{pFedBayes: Personalized Bayesian Federated Learning Algorithm}
	\begin{tabular}{l} \label{Algorithm_PerFedBayes}
{\bf{Server executes:}} \\
	\hspace{0.4cm}\bf{Input} $T, R, S, \beta, a, b, \bm v^{0}=(\bm \mu^0,\bm\rho^0)$ \\
	\hspace{0.4cm}\bf{for} $t=0,1,...,T-1$ \bf{do} \\
      	\hspace{0.8cm}\bf{for} $i=1,2,...,N$ \bf{in parallel do} \\
			\hspace{1.5cm}$\bm v_i^{t+1} \leftarrow \text{ClientUpdate}(i,\bm v^t)$  \\
		\hspace{0.8cm}$\mathbb{S}^{t} \leftarrow \text{Random index sets of $S$ clients}$ \\
		\hspace{0.8cm}$\bm v^{t+1}=(1-\beta) \bm v^{t}+\frac{\beta}{S} \sum_{i \in \mathbb{S}^{t}} {\bm v_i^{t+1}}$
	\\
	$\textbf{ClientUpdate}(i,\bm v^{t})\textbf{:}$ \\
		\hspace{0.4cm}$ \bm v_{w,0}^{t} = \bm v^{t}$  \\
		    \hspace{0.4cm}{\bf{for}} $r=0,1,...,R-1$ \bf{do}  \\
		        \hspace{0.8cm} $\bm{D}^{i}_\Lambda$ $\leftarrow$ Minibatch sampling with size $b$ from $\bm{D}^i$ \\
		        \hspace{0.8cm} $\bm{g}_{i,r}$ $\leftarrow$ Random sampling with size $a$ from $\mathcal{N}(0,1)$\\
		        \hspace{0.8cm} $\Omega^i(\bm v_{r}^t)$ $\leftarrow$  Apply  (\ref{Client-pFedBayes-2}) with $\bm{g}_{i,r}$, $\bm{D}^{i}_\Lambda$ and $\bm v_{r}^t$\\
		         \hspace{0.8cm} $\bm v_{r}^t$ $\leftarrow$  Update  with $\nabla \Omega(\bm v_{r}^t)$ via GD algorithms \\
		         \hspace{0.8cm}  $\bm v_{w,r+1}^t$ $\leftarrow$ Update with $\nabla \Omega(\bm v_{w,r}^t)$ via GD algorithms \\
		\hspace{0.4cm} return $\bm{v}_{w,R}^{t}$ to the server
	\end{tabular}
\end{minipage}
\end{algorithm}

\subsection{Algorithm}

This subsection presents the implementation of pFedBayes using the stochastic gradient descent (SGD) algorithms, which is shown in Algorithm \ref{Algorithm_PerFedBayes}. In order to represent the distribution of the network parameter $\bm \theta$, we introduce two new  parameters $\bm{\mu}$ and $\bm{\rho}$, where $\mu_m$ represents the mean and $\sigma_m=\log(1+\exp(\rho_m))$ represents the standard deviation of the random variable ${\theta}_m$. The reason we use $\rho_m$ instead of $\sigma_m$ is to make $\sigma_m$ non-negative. Define the parameter tuple $\bm{v}=(\bm{\mu}, \bm{\rho})$. We can reformulate the vector $\bm{\theta}$ as $\bm{\theta}=h({\bm{v}},\bm{g})$, where
\begin{equation} \label{eq: generation}
    \theta_m=h({v_m}, g_m)=\mu_m+\log(1+\exp(\rho_m))\cdot g_m, ~~g_m \sim \mathcal{N}(0,1).
\end{equation}
Then we rewrite the distribution of $\bm{\theta}$ as $q_{\bm v} (\bm{\theta})$, where $q_{\bm v} (\theta_m)= \mathcal{N}\big{(}\mu_m,\log^2(1+\exp(\rho_m))\big{)}, m=1,\ldots,T$.

To reduce the cost of communication, we update a {\em localized global model}\footnote{A {\em localized global model} is a global model trained on the client's side and not uploaded to the server \textcolor{black}{during local training}.} and the local model for $R$ rounds alternatively on the clients' side and then aggregate the localized global models in the server. We start with approximating the objective function of local clients in \eqref{Client-1}. Under Gaussian assumptions, the KL divergence term can be calculated by using \eqref{kl_q_w}. But we cannot obtain a closed-form result for the first term in  \eqref{Client-1}. Instead, Monte Carlo approximation is applied to estimation the first term. To accelerate the convergence, we utilize a minibatch of the samples in the approximation. Particularly, the objective function for the $i$-th client is represented as 
\begin{multline}
\label{Client-pFedBayes-2}
   \Omega^i(\bm{v}) \approx 
   - \frac{n}{b} \frac{1}{a} \sum_{j=1}^{b} \sum_{k=1}^{a}    \log p^i_{h({\bm{v}}, \bm{g}_k)}(\bm{D}^i_j) \\+  \zeta\, \mbox{KL}(q^i_{\bm{v}}(\bm{\theta})||w_{\bm{v}}(\bm{\theta})),
\end{multline}
where $b$ and $a$ are minibatch size and Monte Carlo sample size, respectively. The corresponding objective function of the $i$-th localized global model is given by
\begin{align}
\label{Client-pFedBayes-4}
  \Omega^i_w(\bm{v}) = 
     \mbox{KL}(q^i_{\bm{v}}(\bm{\theta})||w_{\bm{v}}(\bm{\theta})).
\end{align}

The clients make $R$ iterations and send the updated localized global models to the server. At each iteration of the clients, the algorithm updates the personalized models with $\nabla_{\vv} \Omega^i(\bm{v})$ and the localized global models with $\nabla_{\vv} \Omega^i_w(\bm{v})$ alternatively. Considering the silence of some clients in practice, we assume that only a random subset of the clients upload their parameters. The subset is defined as $\mathbb{S}^t \in \{1,\ldots, N\}$ and its size is $S$. After receiving the parameters of the clients in $\mathbb{S}^t$, the server takes the mean of the clients and makes an extrapolation step to speed up the algorithm. The parameter for the extrapolation step is defined as $\beta$.

\section{Sparse Personalized Federated Learning with Bernoulli-Gaussian distribution}

\textcolor{black}{A key challenge of Bayesian neural networks (BNNs) in federated learning lies in their expensive inference procedure. Unlike deterministic neural networks that require only a single forward pass, BNNs must sample network parameters from the learned posterior distribution, perform multiple forward passes, and then average the predictions to obtain reliable results \cite{bai2020efficient,cherief2020convergence}. This stochastic inference mechanism, while providing uncertainty quantification, substantially increases both the storage requirement and the computational cost, which becomes particularly problematic in federated settings with resource-constrained clients. To address this issue, we propose sFedBayes, which introduces sparsity into personalized Bayesian federated learning. By adopting a spike-and-slab prior \cite{cherief2020convergence,bai2020efficient}, sFedBayes prunes a large fraction of redundant parameters, resulting in a more compact model.}

In particular, we use Bernoulli-Gaussian distribution as the prior distribution. Suppose that each entry of the $i$-th client $\mathcal{Q}^i$ follows
\begin{multline}\label{ptheta}
\theta_{i,m}|\gamma_{i,m} \sim \gamma_{i,m}\mathcal{N}(\mu_{i,m}, \sigma^2_{i,m}) + (1-
\gamma_{i,m})\delta_0, \\ \gamma_{i,m} \sim \mbox{Bern}(\lambda_{i,m}), \,m=1,\ldots, T, 
\end{multline}
where $\lambda_{i,m}$ represents the inclusion probability, $\mu_{i,m}$ represents the mean and $\sigma^2_{i,m}$ represents the variance of a Gaussian distribution for $m$-th entry of the $i$-th client and $\delta_0$ represents the Dirac at $0$. 
Suppose that each entry of the server satisfies
\begin{multline}\label{wstar}
\theta_{w,m}|\gamma_{w,m} \sim \gamma_{w,m}\mathcal{N}(\mu_{w,m}, \sigma^2_{w,m}) + (1-
\gamma_{w,m})\delta_0, \\ \gamma_{w,m} \sim \mbox{Bern}(\lambda_{w,m}), \,m=1,\ldots, T, 
\end{multline}
where $\lambda_{w,m}$ represents the inclusion probability, $\mu_{w,m}$ represents the mean and  $\sigma^2_{w,m}$ represents the variance of a Gaussian distribution for $m$-th entry of the server. 

Under the above distribution assumptions \eqref{ptheta} and \eqref{wstar}, we can give the upper bound of KL divergence for the two Bernoulli-Gaussian distributions
\begin{align} \label{kl_q_w_s}
    &\mbox{KL}(q^i(\theta)||w(\theta))) \nonumber\\
    &\le \sum_{m=1}^T \mbox{KL}(q^i(\gamma_m)||w(\gamma_m)) \nonumber\\
    &+\sum_{m=1}^T q^i(\gamma_m=1) \mbox{KL}(\mathcal{N}(\mu_{i,m}, \sigma^2_{i,m})||\mathcal{N}(\mu_{w,m}, \sigma^2_{w,m}))  \nonumber\\
    &=\sum_{m=1}^T \left[ \lambda_{i,m} \log\left( \frac{\lambda_{i,m}}{\lambda_{w,m}}\right) + (1-\lambda_{i,m}) \log\left( \frac{1-\lambda_{i,m}}{1-\lambda_{w,m}}\right)\right]  \nonumber\\
    &+  \sum_{m=1}^{T} \frac{\lambda_{i,m}}{2}
    \left[ \log \left( \frac{\sigma_{w,m}^2}{\sigma_{i,m}^2} \right) + \frac{\sigma_{i,m}^2+ (\mu_{i,m}-\mu_{w,m})^2}{\sigma_{w,m}^2} -1
    \right],
\end{align}
where $q^i(\gamma_m)=\lambda_{i,m}^{\gamma_m}(1-\lambda_{i,m})^{(1-\gamma_m)}$,
and 
$w(\gamma_m)=\lambda_{w,m}^{\gamma_m}(1-\lambda_{w,m})^{(1-\gamma_m)}$.

\subsection{\textcolor{black}{Theoretical} analysis}

The generalization error bound for the sparse case can be obtained by simply replacing the total number of parameters $T$ with an optimal sparsity $s^\star_i$ of each neural network \cite{cherief2020convergence}, which is omitted for brevity.

Before giving the results, we define some parameters and give some assumptions. Let $B>0$ be an absolute constant such that $|\theta_{i,m}|, |\theta_{w,m}|\le B$, and define
\begin{align}
&\bar{\chi}^i_n\triangleq\big((L+1)s^\star_i/{n}\big)\log M + ({s^\star_i}/{n})\log\big(r_0 \sqrt{{n}/{s^\star_i}}\big),\\
&\bar{\xi}^i_n\triangleq \inf_{\bm{\theta} \in \Theta(L, \boldsymbol{r}, s^\star),\|\bm{\theta}\|_\infty\leq B}||f_{\theta}^i-f^i||^2_\infty,\\
&\bar{\varepsilon}^i_n\triangleq \sqrt{\bar{\chi}^i_n}\log^\delta(n).
\end{align}

\begin{assumption} \label{assump: lambda}
For the $i$-th cilent, the hyperparameter $\sigma_i^2>0$ is an absolute constant, and $\lambda_i$ satisfies
$\log(1/\lambda_i) = O\{(L+1)\log M + \log(r_0\sqrt{n/s^\star_i})\}$ and 
$\log(1/(1-\lambda_i)) = O((s^\star_i/T)\{(L+1)\log M + \log(r_0\sqrt{n/s^\star_i})\})$.
\end{assumption}

\begin{assumption} \label{assump: sigma_bound_sparse}
The parameters $r_0,n,M,L$  is large enough such that 
\begin{equation}
    \sigma_{i,n}^2=  \frac{s^\star_i}{8n} A \le B^2.
\end{equation}
where $A$ is defined in Eq. \eqref{eq: A}.
\end{assumption}

\begin{theorem} \label{Thm_main_sparse}
If Assumptions \ref{assump: equal-width},\ref{assump: non-linear}, \ref{assump: lambda} and \ref{assump: sigma_bound_sparse} are true, then the following generalization error upper bound is satisfied with dominating
probability
\begin{multline}
\label{neq: local_upperbound}
  \frac{1}{N} \sum_{i=1}^{N} \int_{\Theta}  d^2(P_{\bm{\theta}}^i, P^i) \hat{q}^i(\bm{\theta}) d\bm{\theta}
    \\ \le  \frac{1}{N} \sum_{i=1}^{N} \Big[C [\bar{\varepsilon}^{i}_{n}]^2
  + C'  \bar{\chi}_n^i +\frac{C''}{\zeta}\bar{\xi}^i_n \Big],
\end{multline} 
where $\zeta \ge 1$ is a balance parameter, $C>0$ is a constants and $C', C''$ are arbitrary diverging sequences. 
\end{theorem}

\textcolor{black}{It is worth noting that pFedBayes can be regarded as a special case of sFedBayes when the sparsity level is set to one, i.e., when all parameters remain active. In this sense, sFedBayes generalizes pFedBayes by introducing a learnable sparsity structure on top of the personalized Bayesian framework. A well-chosen sparsity can strike a balance between estimation error and approximation error. As a result, this translates into a favorable trade-off between inference efficiency and model performance, where moderate sparsity yields comparable accuracy to pFedBayes while substantially reducing inference cost.}

Combining with approximation error from \cite{schmidt2020nonparametric}, it can be shown that the upper bound is minimax rate optimal up to a logarithmic term for the sparse personalized federated learning model.

\subsection{Algorithm}

In order to implement the proposed model, we reparameterize $\bm \theta$
with a tuple $\bm v=(\bm \mu,\bm \rho,\bm \lambda)$. After reparameterization, we obtain $\bm \theta=\phi(\bm v, \bm \gamma, \bm g)$, where
\begin{multline}
    \theta_m= \phi(v_m,\gamma_m, g_m)= \gamma_m [\mu_m+\log(1+\exp(\rho_m))\cdot g_m], \\~~\gamma_m \sim  \mbox{Bern} (\lambda_m),~ g_m \sim \mathcal{N}(0,1).
\end{multline}

Since the discrete distribution of Bernoulli makes it difficult to backpropagate the gradient with respect to $\gamma$, we only use the Bernoulli variable in the forward pass and approximate it by a continuous function in the backward pass. Motivated by the work \cite{bai2020efficient}, we use the Gumbel-softmax approximation here. Particularly, we use $\tilde{\gamma} \sim \mbox{Gumbel-softmax}(\lambda,\tau)$ to approach $\gamma \sim \mbox{Bern}(\lambda)$, where
\begin{multline}
        \widetilde{\gamma}  =(1+\exp(-\eta/ \tau))^{-1}, \\ \eta = \log \frac{\lambda}{1 - \lambda} + \log \frac{u}{1-u}, \quad u \sim \mathcal{U}(0,1).
\end{multline}

Define the localized global model of the $i$-th client
\begin{align}
\label{Client-sFedBayes-4}
   &\Omega^i_w(\bm v) =  \nonumber\\
    &\sum_{m=1}^T \left[ \lambda_{i,m} \log\left( \frac{\lambda_{i,m}}{\lambda_{w,m}}\right) + (1-\lambda_{i,m}) \log\left( \frac{1-\lambda_{i,m}}{1-\lambda_{w,m}}\right)\right]  \nonumber\\
    &+  \sum_{m=1}^{T} \frac{\lambda_{i,m}}{2}
    \left[ \log \left( \frac{\sigma_{w,m}^2}{\sigma_{i,m}^2} \right) + \frac{\sigma_{i,m}^2+ (\mu_{i,m}-\mu_{w,m})^2}{\sigma_{w,m}^2} -1
    \right].
\end{align}
For the local model of the $i$-th client, we use the Monte Carlo estimation
to approximate the local loss. In particular, the stochastic estimator for the $i$-th client is given by
\begin{multline}
\label{Client-FedBayes-2}
   \Omega^i(\bm v) \approx 
   - \frac{n}{b} \frac{1}{a} \sum_{j=1}^{b} \sum_{k=1}^{a}  \log p^i_{\phi(\bm v, {\bm \gamma}_k, {\bm g}_k)}(D^i_j) +  \zeta\, \Omega^i_w(\bm v),
\end{multline}
where $b$ and $a$ are minibatch size and Monte Carlo sample size, respectively. So we can give the algorithm in Algorithm \ref{Algorithm_sFedBayes}.

\begin{algorithm}[!htb] \small

\begin{minipage}{\linewidth}
	\caption{sFedBayes: Sparse Personalized  Bayesian Federated Learning Algorithm}
	\begin{tabular}{l} \label{Algorithm_sFedBayes}
{\bf{Server executes:}} \\
	\hspace{0.4cm}\bf{Input} $T, R, S, \tau, \eta, a, b, \beta, \bm v^{0}=(\bm \mu^0,\bm\rho^0,\bm\lambda^0)$ \\
	\hspace{0.4cm}\bf{for} $t=0,1,...,T-1$ \bf{do} \\
      	\hspace{0.8cm}\bf{for} $i=1,2,...,N$ \bf{in parallel do} \\
			\hspace{1.5cm}$\bm v_i^{t+1} \leftarrow \text{ClientUpdate}(i,\bm v^t)$  \\
		\hspace{0.8cm}$\mathbb{S}^{t} \leftarrow \text{Random index sets of $S$ clients}$\\
		\hspace{0.8cm}$\bm v^{t+1}=(1-\beta) \bm v^{t}+\frac{\beta}{S} \sum_{i \in \mathbb{S}^{t}} {\bm v_i^{t+1}}$ \\
	\\
	$\textbf{ClientUpdate}(i,\bm v^{t})\textbf{:}$ \\
		\hspace{0.4cm}$\bm v_{w,0}^{t} = \bm v_{0}^{t} = \bm v^{t}$  \\
		    \hspace{0.4cm}{\bf{for}} $r=0,1,...,R-1$ \bf{do}  \\
		        \hspace{0.8cm} $\mathcal{D}^{i}_\Lambda$ $\leftarrow$ Minibatch sampling with size $b$ from $\mathcal{D}^i$ \\
		        \hspace{0.8cm} $\epsilon_{i,r}, u_{i,r}$ $\leftarrow$ Random sampling with size $a$ from $\mathcal{N}(0,1)$ \\
		        \hspace{2.5cm}and $\mathcal{U}(0,1)$ \\
		        \hspace{0.8cm} $\Omega^i(\bm v_{r}^t)$ $\leftarrow$  Apply \eqref{Client-FedBayes-2} with $\mathcal{D}^{i}_\Lambda$, $\epsilon_{i,r}, u_{i,r}$ and $v_{r}^t$\\
		         \hspace{0.8cm} $\bm v_{r}^t$ $\leftarrow$  Update  with $\nabla \Omega(\bm v_{r}^t)$ via GD algorithms \\
		         \hspace{0.8cm}  $\bm v_{w,r+1}^t$ $\leftarrow$ Update with $\nabla \Omega(\bm v_{w,r}^t)$ via GD algorithms \\
		\hspace{0.4cm}return $\bm{\bm v}_{w,R}^{t}$ to the server
	\end{tabular}
\end{minipage}
\end{algorithm}

\begin{figure}[!t]
      \centering
      \centerline{\includegraphics[width=0.98\linewidth]{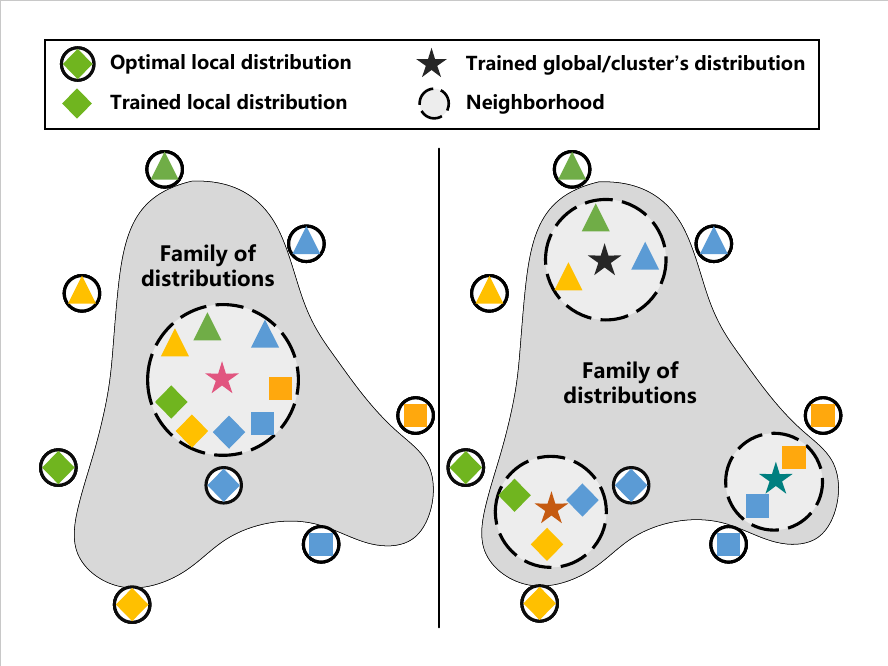}}
\caption{Comparisons for personalized Bayesian federated learning model and clustered Bayesian federated learning model. \textbf{Left:} Personalized Bayesian federated learning model;  \textbf{Right:} Clustered Bayesian federated learning model. The clients with the same shape belong to the same cluster.}
    \label{fig:ClusteredBayesian}
\end{figure}

\section{Clustered Federated Learning via Bayesian Inference}

We propose a clustered Bayesian federated learning model in this section.  When the data among clients are quite different, the global distribution obtained by pFedBayes cannot serve as the good prior distribution. To overcome the extreme heterogeneity from non-i.i.d. data, we consider learning multiple prior distributions to improve the personalization performance. As shown in Fig. \ref{fig:ClusteredBayesian}, the prior distributions generated by clustered FL for different clusters are much better than that generated by vanilla personalized FL, implying that clustered FL can achieve better personalization than vanilla personalized FL.

Let $N$ be the number of clients, and $K$ be the number of clusters with  $K<N$. We propose the following bilevel optimization method named cFedBayes to achieve clustered federated learning via Bayesian inference 
\begin{multline}
	\label{Master-cFedBayes}
	\text {Server:} \min _{\{w^k(\bm{\theta}) \}_{k=1}^{K} }\left\{\frac{1}{N} \sum_{i=1}^{N} F_{i}\big{(}\{w^k\}_{k=1}^{K}\big{)}  \right\},
\end{multline}
\begin{multline}
	\label{Client-cFedBayes}
	\text {Clients:}~\\
	F_i\big{(}\{w^k\}_{k=1}^{K}\big{)}\triangleq \min _{q^i(\bm{\theta}) \in \mathcal{Q}^i, \tilde{k}^i \in [K]}\Big\{- \mathbb{E}_{q^i(\bm{\theta})}[\log p^i(\bm{D}^i|{\bm{\theta}})] \\
	+ \zeta \mbox{KL}(q^i(\bm{\theta})||w^{\tilde{k}^i}(\bm{\theta})) 
    \Big\},
\end{multline}
where $w^k(\bm{\theta})$ is the parameter to be optimized for the $k$-th cluster, $q^i(\bm{\theta})$ is the parameter to be optimized for the $i$-th client and $\tilde{k}^i$ represents the cluster of $i$-th client. Different from pFedBayes, the incorporation of clusters makes  cFedBayes applicable for extreme heterogeneity from non-i.i.d. data.

\subsection{Theoretical analysis}

In order to give an upper bound of the average generalization error, we incorporate a constant in the bilevel optimization problem and the problem of the clients is reformulated as
\begin{multline}
	\label{Client-l}
    F_i\big{(}\{w^k\}_{k=1}^K \big{)} = \min_{q^i(\bm{\theta}) \in \mathcal{Q}^i, \tilde{k}^i \in [K] }\bigg\{\int_{\Theta} l_n(P^{i},P^{i}_{\bm{\theta}}) q^i(\bm{\theta}) d\bm{\theta} \\
	\\+ \zeta  \mbox{KL}(q^i(\bm{\theta})||w^{\tilde{k}^i}(\bm{\theta})) 
 \bigg\},  
\end{multline}
where $l_n(P^{i},P^{i}_{\bm{\theta}})$ is defined in \eqref{def: l_n}.

Let $\{\hat{w}^k(\bm{\theta})\}_{k=1}^K$ be solution of the problem \eqref{Master-cFedBayes} and $\hat{q}^i(\bm{\theta})$ and $\hat{k}^i$ be its corresponding solution for the $i$-th client's subproblem
\begin{multline} \label{prb: subproblem_optimal}
	(\hat{q}^i(\bm{\theta}), \hat{k}^i )
	=\arg \min_{q^i(\bm{\theta}) \in \mathcal{Q}^i, \tilde{k}^i \in [K] }\bigg\{\int_{\Theta} l_n(P^{i},P^{i}_{\bm{\theta}}) q^i(\bm{\theta}) d\bm{\theta} \\
	+ \zeta \mbox{KL}(q^i(\bm{\theta})||\hat{w}^{\tilde{k}^i}(\bm{\theta})) 
     \bigg\}.
\end{multline}
Here, we suppose that only one cluster minimizes $\int_{\Theta} l_n(P^{i},P^{i}_{\bm{\theta}}) \hat{w}^k(\bm{\theta}) d\bm{\theta}$; otherwise, we can choose one index randomly that minimizes $\int_{\Theta} l_n(P^{i},P^{i}_{\bm{\theta}}) \hat{w}^k(\bm{\theta}) d\bm{\theta}$.

Let $\mathbb{S}_k=\{i|\hat{k}^i=k\}$ denote the index set of all clients that belong to the $k$-th cluster. Our goal is to give an upper bound for each cluster
\begin{align} \label{dist}
	\frac{1}{|\mathbb{S}_k|}  \sum_{i\in \mathbb{S}_k} \int_{\Theta}   d^2(P_{\bm{\theta}}^i, P^{i}) \hat{q}^i(\bm{\theta}) d\bm{\theta},~ k=1,\ldots,K. 
\end{align}
with parameters $\chi_n$, $\xi^i_n$ and $\varepsilon_n$.

We first give the upper bound of $\int_{\Theta}   d^2(P_{\bm{\theta}}^i, P^{i}) \hat{q}^i(\bm{\theta}) d\bm{\theta}$ in Lemma \ref{lm: clustered upper bound 1} and then give the upper bound of the right-hand side of Lemma \ref{lm: clustered upper bound 1} in Lemma \ref{lm: clustered upper bound 2}.

\begin{lemma} \label{lm: clustered upper bound 1}
If Assumptions \ref{assump: equal-width} and \ref{assump: non-linear} hold, then the following inequality is satisfied with dominating probability
	\begin{multline} \label{neq: clustered local_upperbound_sum}
	  \int_{\Theta}  d^2(P_{\bm{\theta}}^i, P^i)  \hat{q}^i(\bm{\theta}) d\bm{\theta} \le 
		\frac{1}{n} \Bigg\{ \frac{1}{\zeta} \int_{\Theta} l_n(P^{i}, P^{i}_{\bm{\theta}})\hat{q}^i(\bm{\theta}) d\bm{\theta}  \\ + \mbox{\rm KL}(\hat{q}^i(\bm{\theta})||\hat{w}^{\hat{k}^i}(\bm{\theta})) \Bigg\} + C\varepsilon^{2}_{n}, 
	\end{multline}
where $\zeta \ge 1$ is a balance parameter and $C>0$ is a constant.
\end{lemma}

\begin{lemma}
\label{lm: clustered upper bound 2}
If Assumptions \ref{assump: equal-width}-\ref{assump: sigma_bound} hold, then the following inequality is satisfied with dominating probability
\begin{multline} \label{eq: clustered upper bound 2}
    \frac{1}{|\mathbb{S}_k|} \sum_{i\in \mathbb{S}_k}   \Bigg[  \int_{\Theta} l_n(P^{i},P^{i}_{\bm{\theta}}) \hat{q}^i(\bm{\theta}) d\bm{\theta} 
    +  \zeta\, \mbox{\rm KL}(\hat{q}^i(\bm{\theta})||\hat{w}^{\hat{k}^i}(\bm{\theta})) \Bigg] 
    \\ \le n\left(  C' \zeta \chi_n +\frac{C''}{|\mathbb{S}_k|} \sum_{i\in \mathbb{S}_k} \xi^i_n \right),
 \end{multline}
 where $\zeta \ge 1$ is a balance parameter and $C', C''$ are arbitrary diverging sequences.
\end{lemma}

By combining Lemmas \ref{lm: clustered upper bound 1} and \ref{lm: clustered upper bound 2}, we obtain the generalization error bound for cFedBayes as follows.

\begin{theorem} \label{Thm_main_cluster}
If Assumptions \ref{assump: equal-width}-\ref{assump: sigma_bound} hold, then the following generalization error upper bound  is satisfied with dominating probability
\begin{multline}
\label{neq: cluster_local_upperbound}
  \frac{1}{|\mathbb{S}_k|} \sum_{i\in \mathbb{S}_k} \int_{\Theta}  d^2(P_{\bm{\theta}}^i, P^i) \hat{q}^i(\bm{\theta}) d\bm{\theta}
    \\ \le  C \varepsilon^{2}_{n}
  + C'  \chi_n +\frac{C''}{|\mathbb{S}_k|\zeta} \sum_{i\in \mathbb{S}_k}\xi^i_n,
\end{multline}
where $\zeta \ge 1$ is a  balance parameter, $C>0$ is an constant and $C', C''$ are arbitrary diverging sequences. 
\end{theorem}

Suppose that the unknown functions $\{f^i\}_{i \in \mathbb{S}_k}$ are $\beta_k$-Hölder-smooth and all data has $d$ intrinsic dimension. In this case, the order of the generalization error rate for pFedBayes  appears as
\begin{multline} \label{eq: pensonalized_convegencerate}
    	\frac{1}{N}  \sum_{i=1}^N \int_{\Theta}  d^2(P_{\bm{\theta}}^i, P^i) \hat{q}^i(\bm{\theta}) d\bm{\theta} \\= \mathcal{O}\left( \frac{1}{N}\sum_{k=1}^K |\mathbb{S}_k| n^{-\frac{2\beta_k}{2\beta_k+d}}\log^{2\delta'} (n)\right) \\
        = \mathcal{O}\left(  n^{-\frac{2\beta}{2\beta+d}}\log^{2\delta'} (n) \right),
\end{multline}
where $\beta=\min\{\beta_1,\ldots,\beta_K\}$. Following the discussions from the minimax optimality of pFedBayes, we conclude that the convergence rate of the generalization error of cFedBayes is minimax optimal up to a logarithmic term for bounded $\{f^i\}_{i \in \mathbb{S}_k}$ and $\{f^i_{\bm \theta}\}_{i \in \mathbb{S}_k}$. And the order of the generalization error rate for cFedBayes  is 
\begin{multline} \label{eq: cluster_convegencerate}
    	\frac{1}{|\mathbb{S}_k|}  \sum_{i\in \mathbb{S}_k} \int_{\Theta}  d^2(P_{\bm{\theta}}^i, P^i) \hat{q}^i(\bm{\theta}) d\bm{\theta} = \mathcal{O}\left(n^{-\frac{2\beta_k}{2\beta_k+d}}\log^{2\delta'} (n)\right)
\end{multline}
for $k=1,\ldots,K$. 

{ \color{black}
Eq. \eqref{eq: pensonalized_convegencerate} provides a generalization bound for the overall averaged error across all clients. Since it involves a summation over clusters, its asymptotic order is governed by the slowest-decaying term, 
corresponding to the smallest smoothness parameter among all clusters. In contrast, Eq. \eqref{eq: cluster_convegencerate} establishes a cluster-wise generalization bound. 
For each cluster, the rate depends only on its own smoothness parameter. 
Thus, cFedBayes provides cluster-level adaptive guarantees rather than a single uniform bound across all clients.
}

\subsection{Algorithm}
This section describes how to integrate cFedBayes in a federated learning system as shown in Algorithm 
\ref{Algorithm_cFedBayes}. To reduce the communication cost, the global parameters are updated for a few rounds on the clients' side and then aggregated on the server's side. 

Define $ \bm{v}=(\bm{\mu}, \bm{\rho})$. Then $\bm{\theta}$ can be reparameterized as
$\bm{\theta}=h({\bm{v}},\bm{g})$, where
\begin{equation} \label{eq: cgeneration}
	\theta_m=h({v_m}, g_m)=\mu_m+\log(1+\rho_m)\cdot g_m, ~~g_m \sim \mathcal{N}(0,1).
\end{equation}
For any $q \sim \{\mathcal{Q}^i\}_{i=1}^{N} \cup \mathcal{Q}^w$, the distribution of the random vector $\bm{\theta}$ is rewritten as $q_{\bm v} (\bm{\theta})$, where $q_{\bm v} (\theta_m)= \mathcal{N}(\mu_m,\rho_m^2), m=1,\ldots,T$.

\begin{algorithm}[htb] \small
	
	\begin{minipage}{\linewidth}
		\caption{cFedBayes: Clustered Bayesian Federated Learning Algorithm}
		\begin{tabular}{l} \label{Algorithm_cFedBayes}
			{\bf{Server executes:}} \\
			\hspace{0.4cm}\bf{Input} $T, R, S, \beta, a, b, \bm v^{0}=(\bm \mu^0,\bm\rho^0)$ \\
            \hspace{0.4cm}\textcolor{black}{{\bf{Initialization:}} Perform spectral clustering to obtain $\{\bm u^0_{k}\}_{k=1}^{K}$} \\
			\hspace{0.4cm}\bf{for} $t=0,1,...,T-1$ \bf{do} \\
			\hspace{0.8cm}\bf{for} $i=1,2,...,N$ \bf{in parallel do} \\
			\hspace{1.5cm}$\bm \tilde{k}_{i}^{t+1}, \bm v_{w,i}^{t+1} \leftarrow \text{ClientUpdate}(i,\{\bm u^t_{k}\}_{k=1}^{K})$  \\
			\hspace{0.8cm}\bf{for} $k=1,2,...,K$ \bf{in parallel do} \\
             \hspace{1.5cm}\textcolor{black}{$\mathbb{S}_k^{t+1} \leftarrow \text{Random index set of clients for cluster~} k$}  \\
			\hspace{1.5cm}$\bm u^{t+1}_k=(1-\beta) \bm u^{t}_k+\frac{\beta}{|\mathbb{S}_k^{t+1}|} \underset{i \in \mathbb{S}_k^{t+1}}{\sum} \mathbf{1}(\tilde{k}_{i}^{t+1}=k) {\bm v_{w,i}^{t+1}}$\\
	
			\textbf{ClientUpdate}$(i, \{\bm u^t_{k}\}_{k=1}^{K})\textbf{:}$ \\
			\hspace{0.4cm} $\tilde{k}^{t}_i$ $\leftarrow$   Use \eqref{Client-cFedBayes-ID} with $\{\bm u^t_{k}\}_{k=1}^{K}$ \\ 
			\hspace{0.4cm} $\bm v_{w,0}^{t} = \bm u^{t}_{\tilde{k}_i^{t}}$  \\
			\hspace{0.4cm}{\bf{for}} $r=0,1,...,R-1$ \bf{do}  \\
			\hspace{0.8cm} $\bm{D}^{i}_\Lambda$ $\leftarrow$ Minibatch sampling with size $b$ from $\bm{D}^i$ \\
			\hspace{0.8cm} $\bm{g}_{i,r}$ $\leftarrow$ Random sampling with size $a$ from $\mathcal{N}(0,1)$\\
			\hspace{0.8cm} $\Omega^i(\bm v_{r}^t)$ $\leftarrow$ Apply  (\ref{Client-cFedBayes-2}) with $\bm{g}_{i,r}$, $\bm{D}^{i}_\Lambda$ and $\bm v_{r}^t$\\
			\hspace{0.8cm} $\bm v_{r}^t$ $\leftarrow$ Update with $\nabla_{\bm{v}} \Omega^i(\bm v_{r}^t)$ via GD algorithms \\
			\hspace{0.8cm} $\bm v_{w,r+1}^t$ $\leftarrow$  Update with $\nabla \Omega^i_w(\bm v_{w,r}^t)$ via GD algorithms \\
			\hspace{0.4cm} return $(\tilde{k}_{i}^{t},\bm{v}_{w,R}^{t})$ to the server
		\end{tabular}
	\end{minipage}
\end{algorithm}

The cost function for the $i$-th client is shown as
\begin{multline}
	\label{Client-cFedBayes-2}
	\Omega^i(\bm{v}) \approx 
	- \frac{n}{b} \frac{1}{a} \sum_{j=1}^{b} \sum_{l=1}^{a}    \log p^i_{h({\bm{v}},\bm{g}_l)}(\bm{D}^i_j) \\+  \zeta\, \mbox{KL}(q^i_{\bm{v}}(\bm{\theta})||w^{\tilde{k}^i}_{\bm{v}}(\bm{\theta})), \bm \theta \sim q^i_{\bm{v}},
\end{multline}
where
{\color{black}
\begin{equation} \label{Client-cFedBayes-ID}
    \tilde{k}^i = \underset{k}{\mbox{arg min}} \, \mbox{KL}(q^i_{\bm{v}}(\bm{\theta})||w^{k}_{\bm{v}}(\bm{\theta})).
\end{equation}
}

The objective function of the $i$-th localized global model is represented by
\begin{align}
	\label{Client-cFedBayes-4}
	\Omega^i_w(\bm{v}) = 
	\mbox{KL}(q^i_{\bm{v}}(\bm{\theta})||w^{\tilde{k}^i}_{\bm{v}}(\bm{\theta})).
\end{align}

\textcolor{black}{At the beginning of training, all clients receive the initial global model from the server and perform $R$ local update steps using their private datasets $\bm{D}^i$. After the local optimization, the server collects the updated variational distributions $\{ q^i_{\bm{v}_{w,i}^0}(\bm{\theta})\}_{i=1}^{N}$ from all $N$ clients, where $\bm{v}_{w,i}^0$ denotes the parameters of the $i$-th localized global model, and computes pairwise distances based on the symmetric KL divergence:
\begin{multline}
\bm{\Psi}_{\mathrm{sym}}^{(i,j)} = \tfrac{1}{2} \Big( \mbox{KL}\big(q^i_{\bm{v}^0_{w,i}}(\theta)\,\|\,q^j_{\bm{v}^0_{w,j}}(\theta)\big) \\
+ \mbox{KL}\big(q^j_{\bm{v}^0_{w,i}}(\theta)\,\|\,q^i_{\bm{v}^0_{w,j}}(\theta)\big) \Big).
\end{multline}
The resulting divergence matrix $\bm{\Psi}_{\mathrm{sym}} \in \mathbb{R}^{N\times N}$ is converted into a similarity matrix $\bm{\Phi}  \in \mathbb{R}^{N\times N}$ by an inverse-distance kernel, e.g.,
\begin{equation}
\bm{\Phi}_{ij} = \frac{1}{\bm{\Psi}_{\mathrm{sym}}^{(i,j)} + \iota},
\end{equation}
where $\iota>0$ is a small constant to ensure numerical stability. Spectral clustering is then applied to $\bm{\Phi}$ to partition clients into $K$ clusters \cite{ng2001spectral,von2007tutorial} and obtain the initial cluster parameters $\{\bm u^0_{k}\}_{k=1}^{K}$. The resulting cluster centroids are then used to initialize the cluster-specific priors and client memberships. 
}

On the clients' side, the clients first update the cluster identity using Eq. \eqref{Client-cFedBayes-ID} and then make some local iterations to update their parameters and localized global parameters. Then the clients upload their cluster identity and localized global parameters to the server. Based on the uploaded information, the server updates the parameters for all clusters. We can give the algorithm in Algorithm \ref{Algorithm_sFedBayes}.

\begin{table*}[htb]
\caption{Personalization Results on MNIST, FMNIST, CIFAR-10 \textcolor{black}{and SVHN}. Best results are bolded.}
\label{dif-alg-table-personalized}
\centering
\scalebox{1}{
\begin{tabular}{cccccccc}
\toprule
\multirow{3}{*}{Dataset}  & \multirow{3}{*}{Method} & \multicolumn{2}{c}{Small (Acc. ($\%$))}  & \multicolumn{2}{c}{Medium (Acc. ($\%$))}  & \multicolumn{2}{c}{Large (Acc. ($\%$))}\\
\cmidrule(r){3-4}
\cmidrule(r){5-6}
\cmidrule(r){7-8}
&   & {PM} & {GM} & {PM} & {GM} & {PM} & {GM}\\
\midrule

\multirow{10}{*}{MNIST} 
& FedAvg \cite{mcmahan2017communication} & - & 87.38±0.27 & - & 90.60±0.19 & - & 92.39±0.24 \\
& Fedprox \cite{li2018federated} & - & 87.65±0.30 & - & 90.66±0.17 & - & 92.42±0.23 \\
& BNFed \cite{yurochkin2019bayesian} & - & 78.70±0.69 & - & 80.02±0.60 & - & 82.95±0.22 \\
& FedPA \cite{al2021federated} & - & 89.83±0.91 & - & 92.31±0.37 & - & 94.08±0.35 \\
& Per-FedAvg \cite{fallah2020personalized} & 89.29±0.59 & - & 95.19±0.33 & - & 98.27±0.08 & - \\
& pFedMe \cite{t2020personalized} & 92.88±0.04 & 87.35±0.08 & 95.31±0.17 & 90.60±0.19 & 96.42±0.08 & 91.25±0.14 \\
& HeurFedAMP \cite{huang2021personalized} & 90.89±0.17 & - & 94.74±0.07 & - & 96.90±0.12 & - \\
& pFedGP \cite{achituve2021personalized} & 85.96±2.30 & - & 91.96±0.97 & - & 95.66±0.43 & - \\
&  pFedBayes (400) & 
\textbf{94.72±0.18} & 90.16±0.27 &
{97.41±0.03} & 92.58±0.09 &
98.49±0.14 & 94.24±0.45 \\
&  pFedBayes (800) & {94.46±0.15} & \textbf{90.53±0.24} & \textbf{97.43±0.02} & \textbf{92.61±0.06} & \textbf{98.81±0.11} & \textbf{94.41±0.42} \\

\midrule

\multirow{10}{*}{FMNIST} 
& FedAvg \cite{mcmahan2017communication} & - & 81.51±0.19 & - & 83.90±0.13 & - & \textbf{85.42±0.14} \\
& Fedprox \cite{li2018federated} & - & \textbf{81.53±0.08} & - & \textbf{83.92±0.21} & - & 85.32±0.14 \\
& BNFed \cite{yurochkin2019bayesian} & - & 66.54±0.64 & - & 69.68±0.39 & - & 70.10±0.24 \\
& FedPA \cite{al2021federated} & - & 81.06±0.18 & - & 83.71±0.43 & - & 84.60±0.29 \\
& Per-FedAvg \cite{fallah2020personalized} & 79.79±0.83 & - & 84.90±0.47 & - & 88.51±0.28 & - \\
& pFedMe \cite{t2020personalized} & 88.63±0.07 & 81.06±0.14 & 91.32±0.08 & 83.45±0.21 & 92.02±0.07 & 84.41±0.08 \\
& HeurFedAMP \cite{huang2021personalized} & 86.38±0.24 & - & 89.82±0.16 & - & 92.17±0.12 & - \\
& pFedGP \cite{achituve2021personalized} & 86.99±0.41 & - & 90.53±0.35 & - & 92.22±0.13 & - \\
&  pFedBayes (400) & 
89.55±0.07 & 79.45±0.64 &
\textbf{91.98±0.05} & 82.46±0.22 &
92.44±0.03 & 82.93±0.19 \\
& pFedBayes (800) &  \textbf{89.63±0.08} &  78.73±0.62 &  {91.96±0.04} &  82.43±0.15 &  \textbf{93.05±0.04} &  83.41±0.14 \\

\midrule

\multirow{10}{*}{CIFAR10}
& FedAvg \cite{mcmahan2017communication} & - & 44.24±3.01 & - & 56.73±1.81 & - & \textbf{79.05±0.44} \\
& Fedprox \cite{li2018federated} & - & 43.70±1.38 & - & 57.35±3.11 & - & 77.65±1.62 \\
& BNFed \cite{yurochkin2019bayesian} & - & 34.00±0.16 & - & 39.52±0.56 & - & 44.37±0.19 \\
& FedPA \cite{al2021federated} & - & 45.99±0.97 & - & 58.26±1.53 & - & 68.27±1.40 \\
& Per-FedAvg \cite{fallah2020personalized} & 33.96±1.12 & - & 52.98±1.21 & - & 69.61±1.21 & - \\
& pFedMe \cite{t2020personalized} & 49.66±1.53 & 43.67±2.14 & 66.75±1.84 & 51.18±2.57 & 77.13±1.06 & 70.86±1.04 \\
& HeurFedAMP \cite{huang2021personalized} & 46.72±0.39 & - & 59.94±1.42 & - & 73.24±0.80 & - \\
& pFedGP \cite{achituve2021personalized} & 43.66±0.32 & - & 58.54±0.40 & - & 72.45±0.19 & - \\
& pFedBayes (400) & 
60.51±1.35 & 45.53±2.81 &
\textbf{75.58±0.20} & 59.75±3.32 &
\textbf{83.74±0.23} & 62.42±4.07 \\
& pFedBayes (800) &  \textbf{61.63±1.45} &  \textbf{47.90±1.25} &  {75.21±0.39} &  \textbf{61.22±2.35} &  {83.52±0.11} &  64.12±2.15 \\

\midrule

\multirow{7}{*}{SVHN}
& FedAvg \cite{mcmahan2017communication} & - & {53.81±0.03} & - & 81.94±0.08 & - & \textbf{91.51±0.04} \\
& Fedprox \cite{li2018federated} & - & \textbf{53.91±0.04 }& - & 81.12±0.06 & - & 91.17±0.05 \\
& Per-FedAvg \cite{fallah2020personalized} & 56.79±0.07 & - & 88.67±0.08 & - & 92.66±0.01 & - \\
& pFedMe \cite{t2020personalized} & 51.81±0.03 & 42.93±0.04 & 84.30±0.01 & 79.23±0.01 & 88.77±0.04 & 81.02±0.03 \\
& HeurFedAMP \cite{huang2021personalized} & - & 53.62±0.34 & - & 80.81±0.05 & - &  87.82±0.13 \\
& pFedBayes (400) & \textbf{62.53±0.19} & 50.90±0.72 & \textbf{89.70±0.36} & 87.62±0.26 & \textbf{92.84±0.37} & 88.25±0.24 \\
& pFedBayes (800) & {62.48±0.22} & 51.15±0.68 & {89.63±0.31} & \textbf{87.85±0.29} & {92.79±0.34} & 88.41±0.28 \\

\bottomrule
\end{tabular}}
\end{table*}

\begin{figure*}[!ht]
    \centering
    \vspace{-12pt}
    \subfloat{\includegraphics[width=0.3\linewidth]{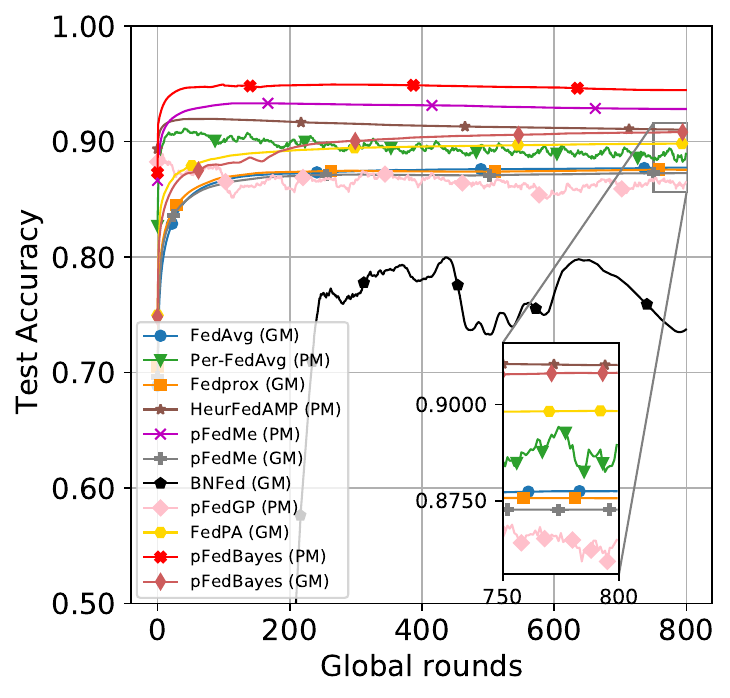}}
    \subfloat{\includegraphics[width=0.3\linewidth]{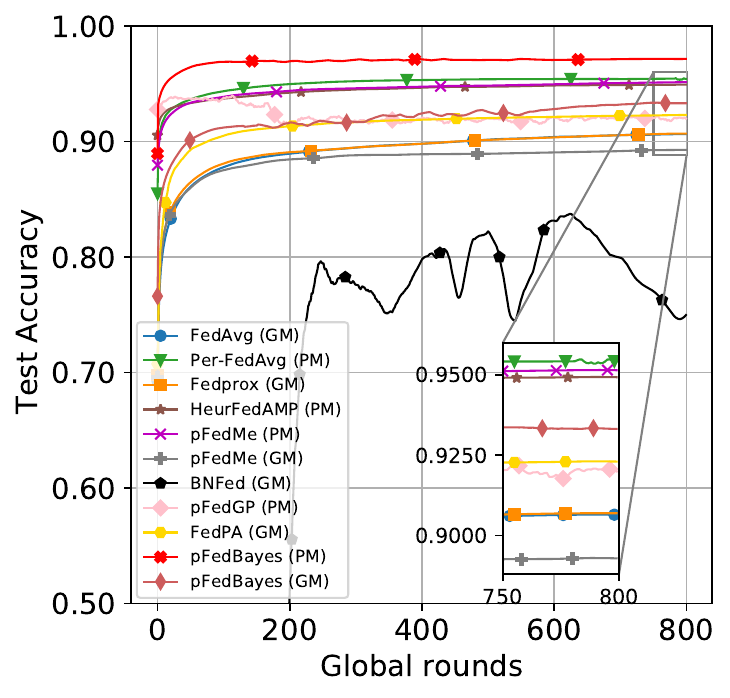}}
    \subfloat{\includegraphics[width=0.3\linewidth]{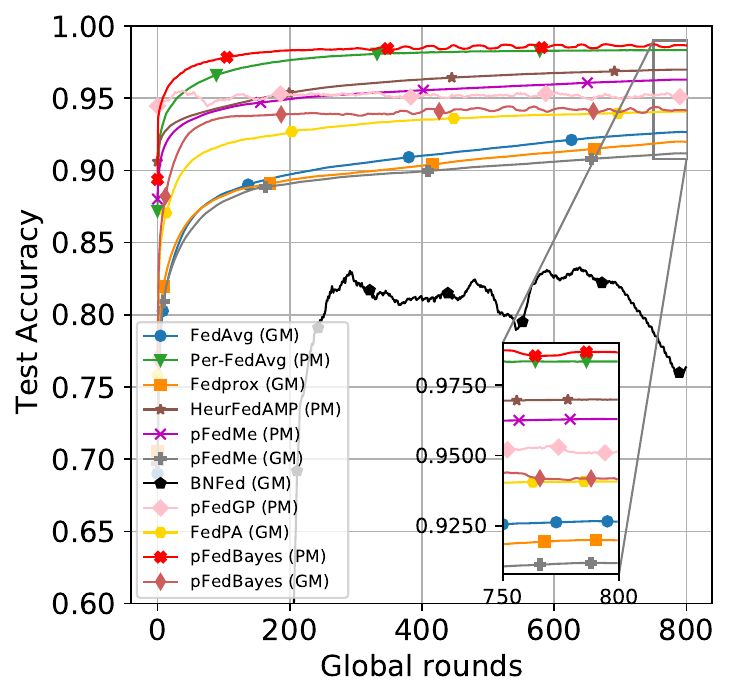}}

    \caption{The convergence curve of the personalization algorithms on the MNIST dataset. \textbf{Left:} Small dataset. \textbf{Middle:} Medium dataset. \textbf{Right:}  Large dataset.}
    \label{fig:fig3}
\end{figure*}

\section{Experiments}
\subsection{Experimental Setting}\label{sec_set}

We validate the proposed pFedBayes with FedAvg \cite{mcmahan2017communication}, Fedprox \cite{li2018federated}, BNFed \cite{yurochkin2019bayesian}, FedPA \cite{al2021federated}, Per-FedAvg \cite{fallah2020personalized}, pFedMe \cite{t2020personalized}, HeurFedAMP \cite{huang2021personalized} and pFedGP \cite{achituve2021personalized} based on non-i.i.d. and clustering datasets. And the proposed cFedBayes is compared with \textcolor{black}{standard federated learning baselines (FedAvg, FedProx, and pFedMe) as well as clustering-based methods including FedSEM~\cite{xie2020multi}, FedCFL~\cite{sattler2021clustered}, and IFCA~\cite{ghosh2020efficient}.}
\textcolor{black}{Four public benchmark datasets, MNIST \cite{lecun2010mnist,lecun1998gradient}, FMNIST (Fashion-MNIST)\cite{xiao2017fashion}, CIFAR-10 \cite{krizhevsky2009learning} and SVHN \cite{netzer2011svhn}  are used to generate the datasets. }
For the MNIST/FMNIST dataset, we used a total of 10 clients, while for the CIFAR-10 dataset, we set up 20 clients. To modify the dataset to be non-i.i.d., a relatively simple approach is to have each client only have a subset of all labels. Here we let each client have only 5 of 10 labels.
For clustering datasets, we construct four types, namely MNIST-MNIST$^{\perp}$, FMNIST-FMNIST$^{\perp}$, MNIST-FMNIST, and CIFAR10-CIFAR10$^{\perp}$. \textcolor{black}{Each dataset contains 10 clients randomly divided into two clusters of equal size.
Here $(\cdot)^{\perp}$ denotes the dataset whose images are rotated by $180^{\circ}$, while the original labels are preserved. 
For MNIST--MNIST$^{\perp}$, FMNIST--FMNIST$^{\perp}$, and CIFAR10--CIFAR10$^{\perp}$, each client holds data from all 10 labels. 
For MNIST--FMNIST, we introduce intra-cluster data heterogeneity by assigning 5 clients with 90\% MNIST and 10\% FMNIST samples, and the other 5 clients with 10\% MNIST and 90\% FMNIST samples.} 
Additionally, we set $S=10$ (random subset of clients) for all experiments. We stop the algorithm at 800 rounds and select the highest accuracy in the last 100 rounds to evaluate the performance of all algorithms.

In addition, in order to verify the performance of the algorithm under different sample sizes, we set three settings of small, medium and large for all datasets. In particular, for MNIST and FMNIST, there are 50 training samples and 950 testing samples per class in the small dataset, 200 training samples and 800 testing samples per class in the medium setting, and 900 training samples and 300 testing samples per class in the large dataset. In contrast, for each class in the CIFAR-10 \textcolor{black}{and SVHN} datasets, we set 25, 100, and 450 training samples and 475, 400, and 150 testing samples for the small, medium, and large settings, respectively.

The computing device we use has two Intel(R) Xeon(R) Gold 6152 CPUs (each with 22 cores, @ 2.10GHz), and two NVIDIA Tesla P100 GPUs with 16GB of memory. The settings of the federated learning environment and the DNN and VGG models are the same as in \cite{t2020personalized}.
Specifically, for the MNIST and FMNIST datasets, we use a DNN model with a hidden layer size of 100. The activation function is ReLU and the network finally passes through the softmax layer.
The VGG model \cite{simonyan2014very} is used for the CIFAR-10 dataset with ``[16, `M', 32, `M', 64, `M', 128, `M', 128, ` M'] " cfg settings. All experiments are implemented based on PyTorch \cite{NEURIPS2019_bdbca288}.

\subsection{Experimental Hyperparameter Settings}

We first study the influence of hyperparameters, and after obtaining the optimal hyperparameters of each algorithm, these algorithms can be compared more fairly. \textcolor{black}{Details of parameter tuning can be found in Appendix B and our conference paper \cite{zhang2022personalized}}. In general, we conduct parameter tuning studies on the medium MNIST dataset, and perform parameter tuning around the optimal parameters recommended in the original papers of these algorithms. For example, we tune the learning rate from 0.0005 to 0.1, because the optimal learning rate recommended by many algorithms (such as pFedMe) is 0.005. For those non-universal hyperparameters, we directly adopt the optimal values recommended in their papers.

\begin{table*}[!t] 
\centering
\caption{Results of sFedBayes \textcolor{black}{under different $\lambda_{\text{init}}$} on MNIST, FMNIST and CIFAR-10. Results better than pFedBayes in Table~1 are bolded. \textcolor{black}{NNR denotes the non-zero ratio.}}
\scalebox{1}{
\begin{tabular}{cccccccc}
\toprule
\textbf{Dataset} & \textbf{Size} & $\boldsymbol{\lambda_{\text{init}}}$ & \textbf{PM Acc.(\%)} & \textbf{PM NNR(\%)} & \textbf{GM Acc.(\%)} & \textbf{GM NNR(\%)} \\
\midrule
\multirow{10}{*}{MNIST} 
& Small  & 0.3 & \textbf{95.43$\pm$0.02} & 20.34$\pm$1.52 & 88.22$\pm$0.38 & 20.45$\pm$1.67 \\
& Small  & 0.5 & \textbf{95.26$\pm$0.15} & 37.67$\pm$1.31 & 89.35$\pm$0.28 & 36.71$\pm$1.91 \\
& Small  & 0.7 & {94.70$\pm$0.15} & 58.21$\pm$0.41 & 88.57$\pm$0.11 & 58.13$\pm$0.59 \\
\cmidrule(lr){2-7}
& Medium & 0.3 & \textbf{97.54$\pm$0.04} & 18.97$\pm$0.47 & 91.12$\pm$0.27 & 17.97$\pm$0.60 \\
& Medium & 0.5 & \textbf{97.60$\pm$0.06} & 35.58$\pm$0.46 & 92.17$\pm$0.11 & 35.40$\pm$0.35 \\
& Medium & 0.7 & 96.93$\pm$0.08 & 66.89$\pm$0.39 & 91.75$\pm$0.37 & 66.66$\pm$0.64 \\
\cmidrule(lr){2-7}
& Large  & 0.3 & \textbf{98.92$\pm$0.02} & 29.44$\pm$0.14 & 93.17$\pm$0.48 & 29.56$\pm$0.07 \\
& Large  & 0.5 & \textbf{98.97$\pm$0.01} & 44.40$\pm$1.00 & 93.39$\pm$0.33 & 42.75$\pm$1.68 \\
& Large  & 0.7 & \textbf{98.94$\pm$0.04} & 55.42$\pm$2.91 & 93.99$\pm$0.29 & 55.46$\pm$1.23 \\
\midrule
\multirow{10}{*}{FMNIST}
& Small  & 0.3 & {89.38$\pm$0.06} & 19.99$\pm$0.72 & 78.85$\pm$0.31 & 19.21$\pm$0.93 \\
& Small  & 0.5 & {89.12$\pm$0.06} & 42.12$\pm$0.62 & 74.61$\pm$0.55 & 42.01$\pm$0.65 \\
& Small  & 0.7 & 88.56$\pm$0.16 & 66.75$\pm$0.54 & 79.08$\pm$0.98 & 66.56$\pm$0.55 \\
\cmidrule(lr){2-7}
& Medium & 0.3 & 91.74$\pm$0.10 & 29.03$\pm$1.68 & 80.07$\pm$2.76 & 29.08$\pm$1.14 \\
& Medium & 0.5 & \textbf{92.00$\pm$0.07} & 47.12$\pm$2.33 & 80.54$\pm$1.46 & 48.93$\pm$0.40 \\
& Medium & 0.7 & \textbf{92.05$\pm$0.08} & 63.92$\pm$0.70 & 80.69$\pm$1.90 & 63.55$\pm$1.67 \\
\cmidrule(lr){2-7}
& Large  & 0.3 & \textbf{93.16$\pm$0.04} & 32.41$\pm$0.03 & 80.46$\pm$2.03 & 32.41$\pm$0.05 \\
& Large  & 0.5 & \textbf{93.33$\pm$0.11} & 47.70$\pm$0.08 & 79.82$\pm$1.16 & 47.71$\pm$0.26 \\
& Large  & 0.7 & \textbf{93.46$\pm$0.11} & 64.84$\pm$0.17 & 81.63$\pm$1.10 & 64.74$\pm$0.19 \\
\midrule
\multirow{10}{*}{CIFAR10}
& Small  & 0.3 & \textbf{63.53$\pm$0.60} & 13.16$\pm$0.29 & 40.48$\pm$0.60 & 13.15$\pm$0.29 \\
& Small  & 0.5 & \textbf{67.47$\pm$0.53} & 26.46$\pm$0.34 & 44.33$\pm$0.74 & 26.54$\pm$0.39 \\
& Small  & 0.7 & \textbf{69.93$\pm$0.69} & 45.79$\pm$1.60 & 47.71$\pm$0.88 & 45.78$\pm$1.58 \\
\cmidrule(lr){2-7}
& Medium & 0.3 & 72.78$\pm$0.86 & 19.59$\pm$0.14 & 48.80$\pm$0.74 & 19.59$\pm$0.13 \\
& Medium & 0.5 & \textbf{77.24$\pm$0.37} & 38.23$\pm$0.13 & 55.04$\pm$0.69 & 38.26$\pm$0.24 \\
& Medium & 0.7 & \textbf{79.61$\pm$0.27} & 58.21$\pm$0.29 & 59.26$\pm$0.30 & 58.23$\pm$0.34 \\
\cmidrule(lr){2-7}
& Large  & 0.3 & 79.92$\pm$0.15 & 26.22$\pm$0.06 & 52.54$\pm$0.52 & 26.17$\pm$0.08 \\
& Large  & 0.5 & 82.65$\pm$0.31 & 44.01$\pm$0.05 & 60.89$\pm$1.23 & 44.04$\pm$0.09 \\
& Large  & 0.7 & \textbf{85.34$\pm$0.37} & 64.41$\pm$0.17 & 68.56$\pm$0.41 & 64.50$\pm$0.16 \\
\bottomrule
\end{tabular}}
\label{dif-sFedBayes-table}
\end{table*}

According to the tuning results, the learning rates of FedAvg, Per-FedAvg, Fedprox, BNFed, pFedGP and HeurFedAMP are set as 0.01, 0.01, 0.01, 0.5, 0.05 and 0.01, respectively. The personalization and global learning rates are both set to 0.01 for both pFedMe and FedPA. The personalization and global learning rates of pFedBayes, sFedBayes, and cFedBayes are all set to 0.001. The personalization regularization weights for Fedprox, pFedMe, and HeurFedAMP are set to 0.001, 15, and $\alpha = 5$, respectively. The personalization regularization weights for pFedBayes, sFedBayes and cFedBayes are all set to $\zeta = 10$. The variance of their weights is initialized to $\rho = -2.5$.

\subsection{pFedBayes Performance Comparison Results}
 
Table \ref{dif-alg-table-personalized} shows the detailed performance comparison results of our pFedBayes algorithm and other personalization algorithms, \textcolor{black}{ where PM denotes the accuracy of the personalized model, GM denotes the accuracy of the global model, and pFedBayes (400) and pFedBayes (800) denote the accuracy after 400 and 800 communication rounds for pFedBayes, respectively.}
We can see that the personalized model of the proposed pFedBayes has significantly improved performance compared to other algorithms, especially on small datasets. \textcolor{black}{From the third column of Table~\ref{dif-alg-table-personalized}, we can see that the personalized model of pFedBayes (800) has improved by 1.69\%, 0.7\%, 12.43\%, and 5.69\% compared with other SOTA algorithms on the small MNIST, FMNIST, CIFAR-10, and SVHN datasets, respectively.} This is because the Bayesian algorithm is more advantageous in small sample conditions, so our pFedBayes converges faster on small datasets. 
Correspondingly, Figure~\ref{fig:fig3} shows convergence curve of the personalization algorithms on the MNIST dataset.
The convergence speed of our pFedBayes is significantly ahead of other algorithms, basically the performance becomes stable after 50 iterations.

In addition, compared with other algorithms, pFedBayes' global model has also achieved a leading position in most settings. This shows that our proposed method can aggregate the distribution of the model very well, which is a capability that previous algorithms did not have. Most of the previous algorithms can only aggregate the weight of the model. Some algorithms that try to aggregate the distribution of the model, such as BNFed, do not perform well.

Furthermore, we can also see that the performance improvement of the proposed algorithm on the FMNIST dataset is relatively small, mainly because the FMNIST dataset is simple and the distribution between different labels is relatively close, the performance of the personalization algorithm has no advantage. A simple and robust algorithm like FedAvg will perform better. 
In contrast, \textcolor{black}{for the complex non-IID CIFAR-10 and SVHN dataset,} the performance improvement of our pFedBayes is significant. Note that on large datasets, our global model does not have much performance advantage. 
Since for large datasets, BNN usually requires some tricks to achieve the same performance as non-Bayesian algorithms~\cite{Osawa2019Practical}. For fair comparison, we did not add special tricks for large datasets to better compare the most essential performance differences between Bayesian federated learning and other algorithms.

\textcolor{black}{For a fair comparison of communication cost, we also report the accuracy of pFedBayes after 400 communication rounds, which approximately matches the number of transmitted bits of FedAvg after 800 rounds. The results show that the performance ranking of different methods remains almost unchanged, confirming that the advantage of pFedBayes is not merely due to additional communication.}

\begin{figure*}[!htb]
    \centering
    \vspace{-12pt}
    \subfloat{\includegraphics[width=0.2\linewidth]{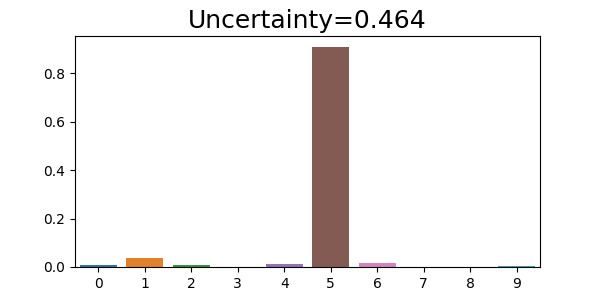}}
    \subfloat{\includegraphics[width=0.2\linewidth]{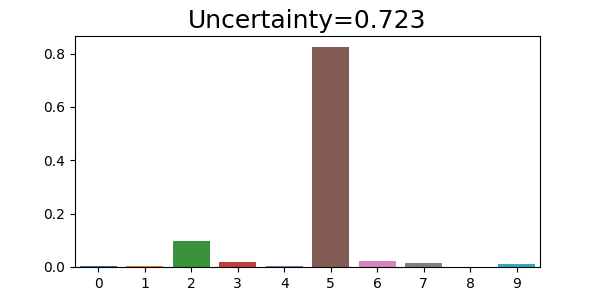}}
    \subfloat{\includegraphics[width=0.2\linewidth]{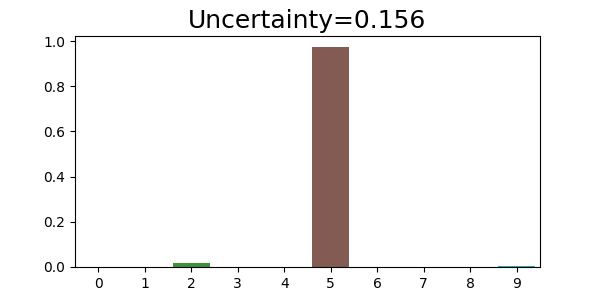}}
    \subfloat{\includegraphics[width=0.2\linewidth]{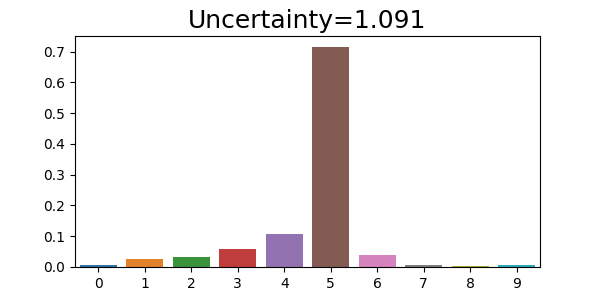}}
    \subfloat{\includegraphics[width=0.2\linewidth]{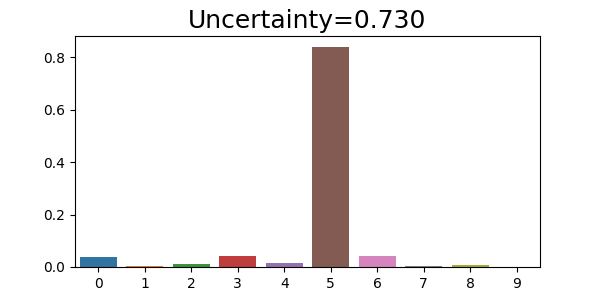}}
    
    \vspace{-12pt}
    \subfloat{\includegraphics[width=0.2\linewidth]{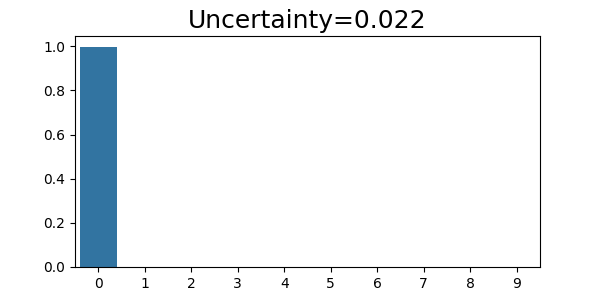}}
    \subfloat{\includegraphics[width=0.2\linewidth]{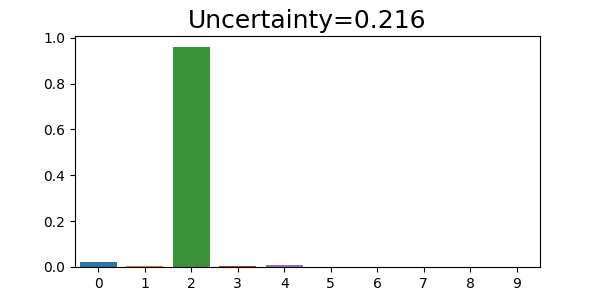}}
    \subfloat{\includegraphics[width=0.2\linewidth]{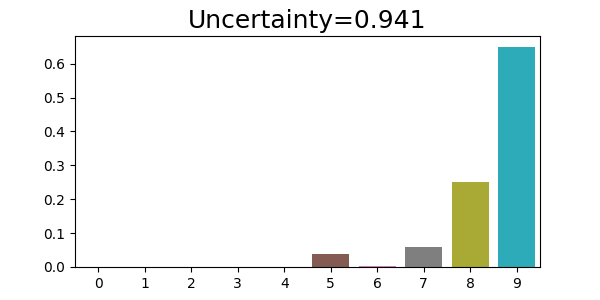}}
    \subfloat{\includegraphics[width=0.2\linewidth]{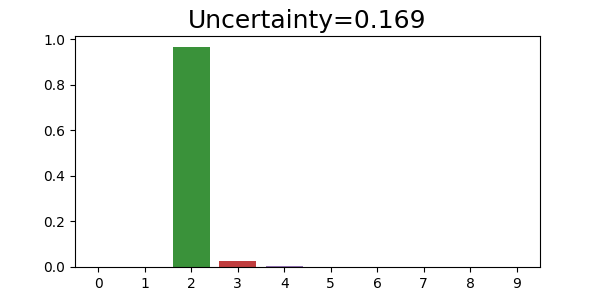}}
    \subfloat{\includegraphics[width=0.2\linewidth]{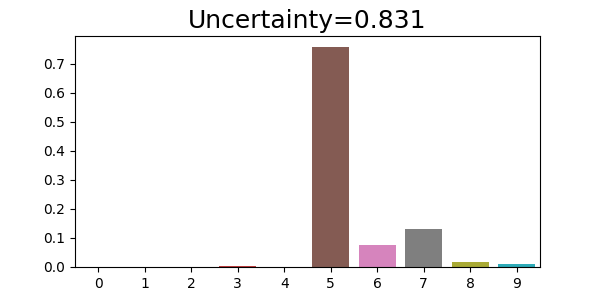}}
    
    \vspace{-12pt}
    \subfloat{\includegraphics[width=0.2\linewidth]{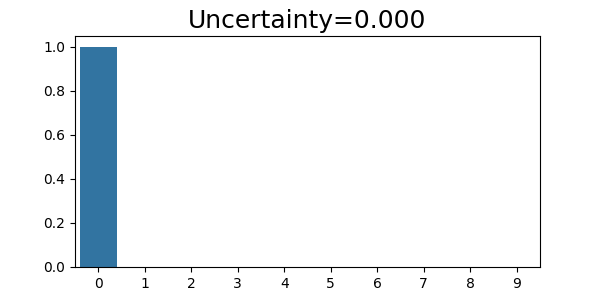}}
    \subfloat{\includegraphics[width=0.2\linewidth]{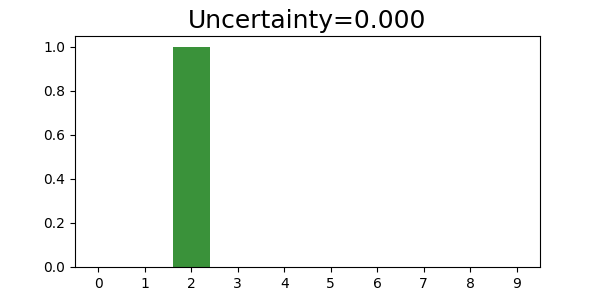}}
    \subfloat{\includegraphics[width=0.2\linewidth]{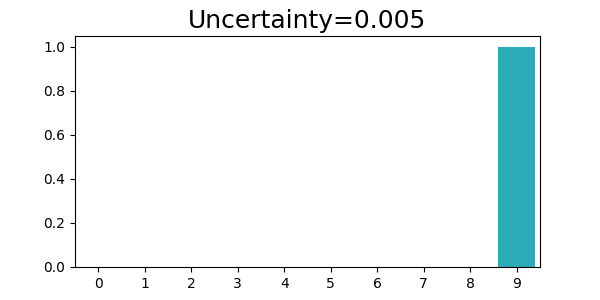}}
    \subfloat{\includegraphics[width=0.2\linewidth]{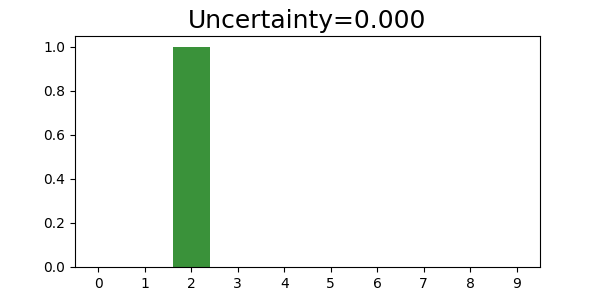}}
    \subfloat{\includegraphics[width=0.2\linewidth]{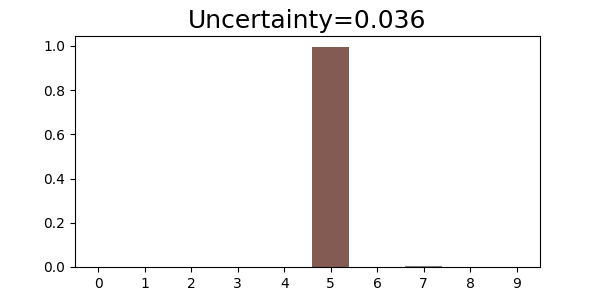}}
    
    \caption{Prediction uncertainty results for the pFedBayes algorithm. We picked five different clients, and the results of training epochs 0, 1, and 10 are given in rows 1, 2, and 3, respectively.}
    \label{fig:fig2}
\end{figure*}

Finally, our pFedBayes can provide results with uncertainty estimates for each client's model, which can help federated learning global models choose which models to aggregate. Specifically, when aggregating models, if we know the uncertainty of all models, we also know the quality of the model, that is, we can decide whether to aggregate the model according to the demand. Figure~\ref{fig:fig2} shows the results of the estimation of prediction uncertainty. For randomly initialized models, each client's uncertainty behaves essentially the same. But as training progresses, each client becomes more confident in its own data classification results. If some models are not confident in their own classification results, we can consider excluding this model during aggregation. Of course, for the sake of fairness, the experiments in this paper did not do so.

\subsection{sFedBayes Performance Comparison Results}

Table~\ref{dif-sFedBayes-table} shows the performance of sFedBayes \textcolor{black}{under different} on small, medium and large datasets of MNIST, FMNIST and CIFAR-10. \textcolor{black}{To numerically verify the sparsity property of the proposed sFedBayes, we report the non-zero ratio (NNR) of the trained Bayesian neural networks under different initial sparsity levels $\lambda_{\text{init}}$. 
As shown in Table~\ref{dif-sFedBayes-table}, the learned sparsity is well aligned with the initialization, confirming that the spike-and-slab prior effectively enforces parameter pruning. 
Interestingly, moderate sparsity, e.g., $\lambda_{\text{init}}=0.3$, $0.5$ or $0.7$, slightly improves the personalized accuracy compared with pFedBayes in Table~\ref{dif-alg-table-personalized}. 
For instance, on MNIST, sFedBayes achieves $97.60\%$ personalized accuracy with only $35.58\%$ of parameters remaining, while on CIFAR-10, $77.24\%$ accuracy is reached with $38.23\%$ non-zero ratio, both surpassing the corresponding pFedBayes results. 
These results demonstrate that sFedBayes can maintain or even improve personalization using less than half of the model parameters, suggesting that the induced sparsity acts as an implicit regularizer that mitigates overfitting. However, this benefit comes at the cost of a slight degradation in the global model accuracy, as the heterogeneous sparsity structure across clients weakens the global parameter aggregation. }

\begin{table*}[!ht]
\caption{\textcolor{black}{Performance comparison on clustering datasets MNIST-MNIST$^{\perp}$, FMNIST-FMNIST$^{\perp}$, CIFAR10-CIFAR10$^{\perp}$ and MNIST-FMNIST. Best results are bolded.}}
\label{dif-alg-table-cluster}
\centering
\scalebox{1}{
\begin{tabular}{cccccccc}
\toprule
\multirow{3}{*}{Dataset}  & \multirow{3}{*}{Method} & \multicolumn{2}{c}{Small (Acc. ($\%$))}  & \multicolumn{2}{c}{Medium (Acc. ($\%$))}  & \multicolumn{2}{c}{Large (Acc. ($\%$))}\\
\cmidrule(r){3-4}\cmidrule(r){5-6}\cmidrule(r){7-8}
&   & {PM} & {GM}& {PM}  &{GM}& {PM}  &{GM}\\
\midrule
\multirow{9}{*}{MNIST-MNIST$^{\perp}$} 
& FedAvg \cite{mcmahan2017communication} & - & {83.08$\pm$0.22} & - & {90.38$\pm$0.06} & - & {92.69$\pm$0.19} \\
& FedProx \cite{li2018federated} & - & {83.32$\pm$0.14} & - & {90.37$\pm$0.04} & - & {92.56$\pm$0.15} \\
& pFedMe \cite{t2020personalized} & {83.78$\pm$0.19} & {80.76$\pm$0.19} & {90.36$\pm$0.23} & {86.14$\pm$0.79} & {92.78$\pm$0.05} & {89.08$\pm$0.27} \\
& FeSEM \cite{xie2020multi} & - & {81.98$\pm$4.33} & - & {88.44$\pm$2.21} & - & {89.88$\pm$0.01} \\
& FedCFL \cite{sattler2021clustered} & - & {84.24$\pm$0.79} & - & {92.05$\pm$1.31} & - & {96.17$\pm$0.13} \\
& IFCA \cite{ghosh2020efficient} & - & {87.55$\pm$0.89} & - & {92.66$\pm$0.54} & - & {95.22$\pm$0.79} \\
& pFedBayes & {86.53$\pm$0.40} & {87.32$\pm$0.20} & {92.89$\pm$0.14} & {92.23$\pm$0.27} & {95.92$\pm$0.02} & {93.61$\pm$0.49} \\
& sFedBayes & {88.12$\pm$0.14} & {87.59$\pm$0.29} & {93.85$\pm$0.05} & {90.47$\pm$0.17} & {96.36$\pm$0.07} & {89.52$\pm$0.30} \\
& cFedBayes & \textbf{89.72$\pm$0.17} & \textbf{89.99$\pm$0.16} & \textbf{94.41$\pm$0.08} & \textbf{94.44$\pm$0.10} & \textbf{96.79$\pm$0.09} & \textbf{96.82$\pm$0.05} \\
\midrule
\multirow{9}{*}{FMNIST-FMNIST$^{\perp}$} 
& FedAvg \cite{mcmahan2017communication} & - & {78.31$\pm$0.10} & - & {83.03$\pm$0.13} & - & {85.46$\pm$0.12} \\
& FedProx \cite{li2018federated} & - & {78.37$\pm$0.10} & - & {83.01$\pm$0.14} & - & {85.33$\pm$0.13} \\
& pFedMe \cite{t2020personalized} & {77.55$\pm$0.03} & {77.86$\pm$0.11} & {82.54$\pm$0.14} & {82.03$\pm$0.16} & {85.18$\pm$0.07} & {84.40$\pm$0.23} \\
& FeSEM \cite{xie2020multi} & - & {77.58$\pm$2.23} & - & {82.36$\pm$0.10} & - & {81.72$\pm$0.96} \\
& FedCFL \cite{sattler2021clustered} & - & {78.40$\pm$1.28} & - & {83.20$\pm$1.04} & - & {86.28$\pm$0.33} \\
& IFCA \cite{ghosh2020efficient} & - & {80.29$\pm$0.83} & - & {84.05$\pm$0.50} & - & {86.34$\pm$0.42} \\
& pFedBayes & {79.28$\pm$0.06} & {79.64$\pm$0.10} & {83.77$\pm$0.13} & {83.80$\pm$0.09} & {85.07$\pm$0.10} & {84.55$\pm$0.44} \\
& sFedBayes & {79.60$\pm$0.06} & {79.98$\pm$0.07} & {83.62$\pm$0.10} & {82.87$\pm$0.07} & {85.85$\pm$0.05} & {82.17$\pm$0.64} \\
& cFedBayes & \textbf{81.61$\pm$0.19} & \textbf{81.64$\pm$0.19} & \textbf{85.14$\pm$0.04} & \textbf{85.27$\pm$0.03} & \textbf{87.27$\pm$0.05} & \textbf{87.54$\pm$0.06} \\
\midrule
\multirow{9}{*}{CIFAR10-CIFAR10$^{\perp}$}
& FedAvg \cite{mcmahan2017communication} & - & {35.94$\pm$0.58} & - & {43.53$\pm$0.52} & - & {58.15$\pm$0.83} \\
& FedProx \cite{li2018federated} & - & {35.71$\pm$0.51} & - & {43.15$\pm$0.69} & - & {57.51$\pm$0.90} \\
& pFedMe \cite{t2020personalized} & {33.86$\pm$0.52} & {30.07$\pm$0.90} & {38.56$\pm$0.67} & {33.19$\pm$0.61} & {54.02$\pm$0.82} & {46.96$\pm$0.19} \\
& FeSEM \cite{xie2020multi} & - & {40.87$\pm$2.87} & - & {49.33$\pm$4.70} & - & {56.55$\pm$6.03} \\
& FedCFL \cite{sattler2021clustered} & - & {39.33$\pm$2.46} & - & {53.20$\pm$0.89} & - & {65.07$\pm$1.83} \\
& IFCA \cite{ghosh2020efficient} & - & {30.27$\pm$2.10} & - & {35.08$\pm$1.38} & - & {44.55$\pm$3.54} \\
& pFedBayes & {41.53$\pm$0.73} & {41.13$\pm$0.50} & {52.88$\pm$0.54} & {50.71$\pm$0.64} & {66.43$\pm$0.42} & {62.95$\pm$0.33} \\
& sFedBayes & \textbf{45.05$\pm$0.71} & {42.94$\pm$0.16} & \textbf{58.24$\pm$0.51} & {55.02$\pm$0.60} & {66.90$\pm$1.12} & {58.32$\pm$5.05} \\
& cFedBayes & {44.03$\pm$0.60} & \textbf{44.06$\pm$0.63} & {55.67$\pm$0.71} & \textbf{55.72$\pm$0.75} & \textbf{68.86$\pm$0.07} & \textbf{68.95$\pm$0.16} \\
\midrule
\multirow{9}{*}{MNIST-FMNIST}
& FedAvg \cite{mcmahan2017communication} & - & {81.41$\pm$0.11} & - & {87.00$\pm$0.19} & - & {89.33$\pm$0.13} \\
& FedProx \cite{li2018federated} & - & {81.49$\pm$0.12} & - & {87.02$\pm$0.21} & - & {89.20$\pm$0.20} \\
& pFedMe \cite{t2020personalized} & {82.95$\pm$0.08} & {80.68$\pm$0.07} & {87.77$\pm$0.21} & {84.63$\pm$0.25} & {90.24$\pm$0.05} & {87.02$\pm$0.08} \\
& FeSEM \cite{xie2020multi} & - & {81.83$\pm$1.83} & - & {86.89$\pm$0.70} & - & {88.58$\pm$0.60} \\
& FedCFL \cite{sattler2021clustered} & - & {79.87$\pm$2.17} & - & \textbf{88.95$\pm$0.57} & - & \textbf{91.72$\pm$0.06} \\
& IFCA \cite{ghosh2020efficient} & - & {63.25$\pm$5.79} & - & {68.85$\pm$9.42} & - & {75.34$\pm$2.71} \\
& pFedBayes & {83.75$\pm$0.30} & {84.13$\pm$0.32} & {90.22$\pm$0.06} & {88.20$\pm$0.22} & {92.41$\pm$0.11} & {90.27$\pm$0.26} \\
& sFedBayes & \textbf{85.60$\pm$0.04} & {84.61$\pm$0.02} & {90.49$\pm$0.06} & {88.56$\pm$0.05} & {92.44$\pm$0.13} & {87.69$\pm$0.16} \\
& cFedBayes & {85.25$\pm$0.48} & \textbf{84.81$\pm$0.10} & \textbf{90.62$\pm$0.05} & {88.77$\pm$0.13} & \textbf{93.22$\pm$0.14} & {90.62$\pm$0.70} \\
\bottomrule
\end{tabular}}
\end{table*}

\subsection{cFedBayes Performance Comparison Results}

{\color{black}
Table~\ref{dif-alg-table-cluster} presents the performance comparison of cFedBayes on the small, medium, and large clustering datasets. Overall, cFedBayes achieves comparable or better results than existing clustered or personalized FL baselines across most datasets.
On the MNIST–MNIST$^{\perp}$ and FMNIST–FMNIST$^{\perp}$ datasets, cFedBayes consistently attains the highest accuracies in both personalized and global models, outperforming the strongest baselines such as FedCFL, IFCA, and pFedBayes by about 1–2\%.
For the more challenging CIFAR10–CIFAR10$^{\perp}$ dataset, cFedBayes improves the global accuracy by approximately 2–3\% compared with the clustering-based baselines FedCFL and FeSEM, while maintaining similar personalized performance to sFedBayes.
In the cross-domain MNIST–FMNIST scenario, cFedBayes performs slightly better or on par with sFedBayes and pFedBayes, suggesting that the clustering mechanism remains effective even when the client data distributions are not strictly separable.

In summary, the updated experiments demonstrate that cFedBayes maintains strong and stable performance across various datasets and heterogeneity levels. Although the gains are moderate in some cases, the improvements are consistent and statistically supported, validating the effectiveness of the proposed category-based clustering mechanism and the robustness of our Bayesian framework. In addition, given the promising results of sFedBayes, we also consider incorporating its sparsity-inducing mechanism into the clustered framework in future work to further enhance the performance and inference efficiency of cFedBayes.
}

\section{Conclusions}

This paper proposes three novel personalized Bayesian federated learning models via variational inference when the data is limited. We first propose pFedBayes to achieve personalization under Gaussian distributions. Then we propose sFedBayes to simplify the network structure. Finally, we propose cFedBayes to address the extreme statistical diversity among clients. The average generalization error bounds are provided for the three approaches, which demonstrate that the proposed methods achieve the minimax optimal convergence rate. Numerical results illustrate that the proposed methods outperform existing advanced personalized methods under non-i.i.d. limited data.

There are several interesting directions for future work. One is to provide the convergence analysis for the proposed methods. One is to study the clustered federated learning model with an unknown number of clusters \textcolor{black}{and explore how it can be integrated with a sparsity-inducing mechanism}. The other is to extend the proposed methods to decentralized settings where only local communications are allowed.



%

\ifCLASSOPTIONcaptionsoff
  \newpage
\fi


\appendices
\setcounter{equation}{0}
\renewcommand{\theequation}{A.\arabic{equation}}

\setcounter{theorem}{0}
\renewcommand{\thetheorem}{A\arabic{theorem}}

\setcounter{lemma}{0}
\renewcommand{\thelemma}{A.\arabic{lemma}}

\section{Proof of Lemmas}
\subsection{Proof of Lemma 1}

Before giving the proof of Lemma 1, we present the following useful lemma for $\mbox{KL}$ divergence \cite{Boucheron2013Concentration}. 
\begin{lemma}{\label{lmdonsker}} Given arbitrary probability measure $\mu$ and arbitrary measurable function $h$ with $e^h \in L_1(\mu)$, we have
	$$
	\log \int e^{h(\eta)}\mu(d\eta) = \sup_{\rho}\left[\int h(\eta)\rho(d \eta) - \mbox{\rm KL}(\rho||\mu)\right].
	$$
\end{lemma}

We start to prove Lemma 1. 
Choosing $\log\eta(P^{i}_{\bm{\theta}}, P^{i}) = l_n(P^{i}_{\bm{\theta}}, P^{i})/\zeta+ n \, d^2(P^{i}_{\bm{\theta}}, P^{i})$ with $\zeta\ge 1$, then we have
\begin{align} \label{lmp: neq1}
	\mathbb{E}_{P^i} \exp(l_n(P^{i}_{\bm{\theta}}, P^{i})/\zeta)&= 
	\mathbb{E}_{P^i} \left(\frac{p^{i}_{\bm{\theta}}(\bm{D}^i)}{p^{i}(\bm{D}^i)} \right)^{\frac{1}{\zeta}} \nonumber\\ 
	&\le  \left(\mathbb{E}_{P^i} \frac{p^{i}_{\bm{\theta}}(\bm{D}^i)}{p^{i}(\bm{D}^i)} \right)^{\frac{1}{\zeta}}=1,
\end{align}
where the inequality follows Jensen's inequality and the concavity of $(\cdot)^{1/\zeta}$.

Combining with the proof from the Theorem 3.1 of \cite{pati2018statistical}, the following inequality holds with high probability
\begin{equation}
	\int_{\Theta}\eta(P^{i}_{\bm{\theta}}, P^{i}) w^\star(\bm{\theta})  d\bm{\theta} \leq e^{C n\varepsilon^{2}_{n}},
\end{equation}
where $C>0$ is a large enough constant. 

Choosing $h(\eta)=\log\eta(P^{i}_{\bm{\theta}},P^i)$, $\mu=w^\star(\bm{\theta})$ and $\rho=\hat{q}^i(\bm{\theta})$ in Lemma \ref{lmdonsker}, we have 
\begin{align} \label{neq: local_upperbound1}
	&\int_{\Theta}  d^2(P_{\bm{\theta}}^i, P^i) \hat{q}^i(\bm{\theta}) d\bm{\theta} \nonumber\\
	\le &  \frac{1}{n}\bigg[ \frac{1}{\zeta}\int_{\Theta} l_n( P^{i}, P^{i}_{\bm{\theta}})\hat{q}^i(\bm{\theta}) d\bm{\theta} +\mbox{KL}(\hat{q}^i(\bm{\theta})||w^\star(\bm{\theta})) \nonumber\\
	&\quad + \log \int_{\Theta} \eta(P^{i}_{\bm{\theta}}, P^{i}) w^\star(\bm{\theta})  d\bm{\theta} \bigg]
	\nonumber\\
	\le&  \frac{1}{n}\left[ \frac{1}{\zeta} \int_{\Theta} l_n( P^{i}, P^{i}_{\bm{\theta}})\hat{q}^i(\bm{\theta}) d\bm{\theta} + \mbox{KL}(\hat{q}^i(\bm{\theta})||w^\star(\bm{\theta})) \right]+ C\varepsilon^{2}_{n}. 
\end{align}

By taking the average of all $N$ clients, we finish the proof: 
\begin{multline} \label{neq: local_upperbound_sum_a}
	\frac{1}{N} \sum_{i=1}^{N} \int_{\Theta}  d^2(P_{\bm{\theta}}^i, P^i) \hat{q}^i(\bm{\theta}) d\bm{\theta} \\\le 
	\frac{1}{n} \bigg\{\frac{1}{N} \sum_{i=1}^{N} \Big[ \frac{1}{\zeta} \int_{\Theta} l_n(P^{i}, P^{i}_{\bm{\theta}})\hat{q}^i(\bm{\theta}) d\bm{\theta} \\+ \mbox{KL}(\hat{q}^i(\bm{\theta})||w^\star(\bm{\theta})) \Big] \bigg\} + C\varepsilon^{2}_{n}.
\end{multline}

\subsection{Proof of Lemma 2}

We begin the proof of Lemma 2 by giving Lemma \ref{lm: optimalsolution}, which shows the relationship of the optimal global distribution  $w(\bm{\theta})$ and the distributions of all clients $\{q^i(\bm{\theta})\}_{i=1}^N$. The proof of Lemma \ref{lm: optimalsolution} is delayed in Section \ref{Proof_Lemma_A2}. 

\begin{lemma} \label{lm: optimalsolution}
	Let $w(\bm{\theta})$ be the solution of the following problem 
	\begin{equation} \label{eq: F_w_m}
		\min_{w(\bm{\theta})} \left\{ F(w)=\frac{1}{N} \sum_{i=1}^{N} \mbox{\rm KL}(q^i(\bm{\theta})||w(\bm{\theta})))\right\}.
	\end{equation}
	Then the following equations hold
	\begin{align}
		\mu_{w,m}&=\frac{1}{N} \sum_{i=1}^{N} \mu_{i,m},\\
		\sigma_{w,m}^2&=\frac{1}{N} \sum_{i=1}^{N} \left[\sigma_{i,m}^2+\left(\mu_{i,m}-\mu_{w,m}\right)^2\right].
	\end{align}
\end{lemma}

Next, we present the proof of Lemma 2. Let $\bm{\theta}^\star_i$ be the minimizer of $\norm{f^i_{\bm{\theta}}-f^i}^2_\infty$ subject to $\norm{\bm{\theta}}_\infty \le B$ and define the local distribution $\tilde{q}^i(\bm{\theta})$ of the $i$-th client:
\begin{equation}\label{vbstar}
	\begin{split}
		& \theta_{i,m} \sim \mathcal{N}(\theta_{i,m}^\star,\sigma^2_{n}), m=1,\dots, T
	\end{split}
\end{equation}
where $\sigma_{n}^2$ follows Assumption 3, i.e., $\sigma_{n}^2=\frac{T}{8n} A$ and 
\begin{multline} \label{eq: A2}
	A=\log^{-1}(3r_0 M)\cdot(2BM)^{-2(L+1)}\bigg[\Big(r_0 + 1 + \frac{1}{BM-1}\Big)^2 \\+ \frac{1}{(2BM)^2-1}
	+\frac{2}{(2BM-1)^2}\bigg]^{-1}.
\end{multline}

Let $\tilde{w}(\bm{\theta})$ represent the solution of
\begin{equation}
	\min_{w(\bm{\theta}) \in \mathcal{Q}}\frac{1}{N} \sum_{i=1}^{N} \mbox{KL}(\tilde{q}^i(\bm{\theta})||w(\bm{\theta}))).
\end{equation}
Lemma \ref{lm: optimalsolution} presents that the distribution $\tilde{w}(\bm{\theta})$ appears as
\begin{equation}\label{wtilde}
	\theta_{\tilde{w},m} \sim \mathcal{N}(\mu_{\tilde{w},m}, \sigma^2_{\tilde{w},m}), m=1,\ldots, T,
\end{equation}
where
\begin{align}
	\mu_{\tilde{w},m}&=\frac{1}{N} \sum_{i=1}^{N} \theta_{i,m}^\star,\\
	\sigma_{\tilde{w},m}^2&=\frac{1}{N} \sum_{i=1}^{N} \left[\sigma_{n}^2+\left(\theta_{i,m}^\star-\mu_{\tilde{w},m}\right)^2\right]. \label{eq: sigma_w_tilde}
\end{align}

Using the optimality of $w^\star(\bm{\theta})$ and $\hat{q}(\bm{\theta})$ implies
\begin{multline}
	\frac{1}{N} \sum_{i=1}^{N}  \left[  \int_{\Theta} l_n(P^{i},P^{i}_{\bm{\theta}}) \hat{q}^i(\bm{\theta}) d\bm{\theta} +  \zeta\, \mbox{KL}(\hat{q}^i(\bm{\theta})||w^\star(\bm{\theta}))\right]
	\\ \le
	\frac{1}{N} \sum_{i=1}^{N}  \left[  \int_{\Theta} l_n(P^{i},P^{i}_{\bm{\theta}}) \tilde{q}^i(\bm{\theta}) d\bm{\theta} +  \zeta\, \mbox{KL}(\tilde{q}^i(\bm{\theta})||\tilde{w}(\bm{\theta}))\right].
\end{multline}

Then we give the upper bound of the right-hand side. Under the assumptions of mean-field decomposition, we have
\begin{align}
	\tilde{q}^i(\bm{\theta}) &= \prod^{T}_{m=1}\mathcal{N}(\theta^\star_{i,m}, \sigma^2_{n}),\\
	\label{eq: sigma_tilde 2}
	\tilde{w}(\bm{\theta}) &=\prod^T_{m=1}\mathcal{N}(\mu_{\tilde{w},m}, \sigma^2_{\tilde{w},m}).
\end{align}
Therefore,
\begin{align} \label{eq:KL_q_w}
	&\mbox{KL}(\tilde{q}^i(\bm{\theta})||\tilde{w}(\bm{\theta})) \nonumber\\
	= &\mbox{KL}\left(\prod^{T}_{m=1}\mathcal{N}(\theta^\star_{i,m}, \sigma^2_{n}) \Bigl|\Bigr| \prod^T_{m=1}\mathcal{N}(\mu_{\tilde{w},m}, \sigma^2_{\tilde{w},m})\right) \nonumber\\
	= &  \sum^T_{m=1}\mbox{KL}\left(\mathcal{N}(\theta^\star_{i,m}, \sigma^2_{n})|| \mathcal{N}(\mu_{\tilde{w},m}, \sigma^2_{\tilde{w},m}) \right) \nonumber\\
	= &  \frac{1}{2}\sum^T_{m=1}\left[\log\Bigl(\frac{\sigma^2_{\tilde{w},m}}{\sigma^2_{n}}\Bigr)+\frac{\sigma^2_{n}+(\theta^\star_{i,m}-\mu_{\tilde{w},m})^{2}}{\sigma^2_{\tilde{w},m}} - 1\right] \nonumber\\
	= & \frac{1}{2}\sum^T_{m=1}\left[\log\Bigl(\frac{\sigma^2_{\tilde{w},m}}{\sigma^2_{n}}\Bigr) - 1\right]+\frac{1}{2} \sum^T_{m=1} \frac{\sigma^2_{n}+(\theta^\star_{i,m}-\mu_{\tilde{w},m})^{2}}{\sigma^2_{\tilde{w},m}} \nonumber\\
	\leq &  \frac{T}{2}\left[\log\left(\frac{\sigma^2_{n}+B^2}{\sigma^2_{n}}\right) - 1\right]+\frac{1}{2} \sum^T_{m=1} \frac{\sigma^2_{n}+(\theta^\star_{i,m}-\mu_{\tilde{w},m})^{2}}{\sigma^2_{\tilde{w},m}},
\end{align}
where the last inequality follows
\begin{equation*}
	\sigma_{\tilde{w},m}^2= \sigma_{n}^2-\mu_{\tilde{w},m}^2+\frac{1}{N} \sum_{i=1}^{N} \theta_{i,m}^{\star2}\le \sigma^2_{n}+B^2.
\end{equation*}

By summing over all clients, we obtain
\begin{align*}
	\frac{1}{N}\sum_{i=1}^{N}\mbox{KL}(\tilde{q}^i(\bm{\theta})||\tilde{w}(\bm{\theta}))
	\le & \frac{1}{N}\sum_{i=1}^{N} \bigg\{ \frac{T}{2}\left[\log\left(\frac{\sigma^2_{n}+B^2}{\sigma^2_{n}}\right) - 1\right] \nonumber\\
	&+\frac{1}{2} \sum^T_{m=1} \frac{\sigma^2_{n}+(\theta^\star_{i,m}-\mu_{\tilde{w},m})^{2}}{\sigma^2_{\tilde{w},m}} \bigg\}\\
	\leq &   \frac{T}{2}\log\left(\frac{\sigma^2_{n}+B^2}{\sigma^2_{n}}\right),
\end{align*}
where the last inequality follows (\ref{eq: sigma_w_tilde}), i.e.,
\begin{equation*}
	\frac{1}{N}\sum_{i=1}^{N} \frac{\sigma^2_{n}+(\theta^\star_{i,m}-\mu_{\tilde{w},m})^{2}}{\sigma^2_{\tilde{w},m}}=1.
\end{equation*}

Using Assumption 3 yields
\begin{align} \label{eq: sum_kl_q_w_2}
	\frac{1}{N}\sum_{i=1}^{N}\mbox{KL}(\tilde{q}^i(\bm{\theta})||\tilde{w}(\bm{\theta})) 
	\leq  \frac{T}{2}  \log \left(\frac{2B^2}{\sigma^2_{n}}\right).
\end{align}

According to the definition of $\sigma_n$ (10), we have
\begin{multline} \label{eq: sum_kl_q_w_2_1}
	\frac{T}{2}\log \left(\frac{2B^2}{\sigma^2_{n}}\right)
	\le T(L+1)\log(2BM)+\frac{T}{2}\log\log(3r_0 M) \\+T\log\left(4r_0 \sqrt{\frac{n}{T}}\right)+ \frac{T}{2} \log(2B^2) \le C' n \chi_n.
\end{multline}
Combining Eqs. \eqref{eq: sum_kl_q_w_2} and \eqref{eq: sum_kl_q_w_2_1} gets
\begin{align} \label{eq: sum_kl_q_w_3}
	\frac{1}{N}\sum_{i=1}^{N}\mbox{KL}(\tilde{q}^i(\bm{\theta})||\tilde{w}(\bm{\theta})) 
	\leq  C' n \chi_n.
\end{align}
Using the technique from the Supplementary of \cite{bai2020efficient}, we obtain
\begin{align}
	\int_{\Theta} l_n(P^{i},P^{i}_{\bm{\theta}}) \tilde{q}^i(\bm{\theta}) d\bm{\theta}
	\le C''(n\chi_n+ n \xi^i_n).
\end{align}
Since $\zeta \ge 1$, we get the final upper bound
\begin{multline}
	\frac{1}{N} \sum_{i=1}^{N}  \left[  \int_{\Theta} l_n(P^{i},P^{i}_{\bm{\theta}}) \hat{q}^i(\bm{\theta}) d\bm{\theta} +  \zeta\, \mbox{KL}(\hat{q}^i(\bm{\theta})||w^\star(\bm{\theta}))\right]  \\\le n\left(  C' \zeta \chi_n +\frac{C''}{N} \sum_{i=1}^{N}\xi^i_n \right).
\end{multline}

\subsection{Proof of Lemma 3}

Choosing $\log\eta(P^{i}_{\bm{\theta}}, P^{i}) = l_n(P^{i}_{\bm{\theta}} P^{i})/\zeta+ n \, d^2(P^{i}_{\bm{\theta}}, P^{i})$ with $\zeta\ge 1$ and Using \eqref{lmp: neq1} gives 
\begin{equation}
	\mathbb{E}_{P^i} \exp(l_n(P^{i}_{\bm{\theta}}, P^{i})/\zeta)\le 1.
\end{equation}
Combining with the proof from the Theorem 3.1 of \cite{pati2018statistical}, we have 
\begin{equation}
	\int_{\Theta}\eta(P^{i}_{\bm{\theta}}, P^{i}) \hat{w}^k(\bm{\theta})  d\bm{\theta} \leq e^{C n\varepsilon^{2}_{n}},\, \mbox{w.h.p.,}
\end{equation}
where $C>0$ is a large number.

By choosing $h(\eta)=\log\eta(P^{i}_{\bm{\theta}},P^i)$, $\mu=\hat{w}^k(\bm{\theta})$ and $\rho=\hat{q}^i(\bm{\theta})$ in Lemma \ref{lmdonsker}, for any $k=1,\ldots,K$, we obtain 
\begin{align} \label{neq: local_upperboundc}
	&\int_{\Theta}  d^2(P_{\bm{\theta}}^i, P^i) \hat{q}^i(\bm{\theta}) d\bm{\theta} \nonumber\\
	\le &  \frac{1}{n}\bigg[ \frac{1}{\zeta}\int_{\Theta} l_n( P^{i}, P^{i}_{\bm{\theta}})\hat{q}^i(\bm{\theta}) d\bm{\theta} +\mbox{KL}(\hat{q}^i(\bm{\theta})||\hat{w}^k(\bm{\theta})) \nonumber\\
	&\quad + \log \int_{\Theta} \eta(P^{i}_{\bm{\theta}}, P^{i}) \hat{w}^k(\bm{\theta})  d\bm{\theta} \bigg]
	\nonumber\\
	\le&  \frac{1}{n}\left[ \frac{1}{\zeta} \int_{\Theta} l_n( P^{i}, P^{i}_{\bm{\theta}})\hat{q}^i(\bm{\theta}) d\bm{\theta} + \mbox{KL}(\hat{q}^i(\bm{\theta})||\hat{w}^k(\bm{\theta})) \right]+ C\varepsilon^{2}_{n}. 
\end{align}
To give a tight upper bound, $k=\hat{k}^i$ is the one that minimizes the KL divergence $\mbox{KL}(\hat{q}^i(\bm{\theta})||\hat{w}^k(\bm{\theta}))$. Therefore, 
\begin{align} \label{neq: local_upperbound_sum}
	&\int_{\Theta}  d^2(P_{\bm{\theta}}^i, P^i)  \hat{q}^i(\bm{\theta}) d\bm{\theta} \nonumber\\
	\le 
	&\frac{1}{n} \Bigg\{ \frac{1}{\zeta} \int_{\Theta} l_n(P^{i}, P^{i}_{\bm{\theta}})\hat{q}^i(\bm{\theta}) d\bm{\theta}   + \mbox{\rm KL}(\hat{q}^i(\bm{\theta})||\hat{w}^{\hat{k}^i}(\bm{\theta}))  \Bigg\} + C\varepsilon^{2}_{n}.
\end{align}

\subsection{Proof of Lemma 4}

Before proving Lemma 4,  we present a corollary of Lemma \ref{lm: optimalsolution} to the optimal parameters of the distribution $w^k(\bm{\theta})$ for fixed distributions of all clients $\{q^i(\bm{\theta})\}_{i \in \mathbb{T}_k}$ in the $k$-th cluster. 

\begin{lemma} \label{lm: clustered optimalsolution}
	Let $\mathbb{T}_k$ be the index set of clients that belong to the $k$-th cluster. Let $w^k(\bm{\theta})$ be the solution of the following problem 
	\begin{equation} \label{eq: F_w_m_c}
		\min_{w^k(\bm{\theta})} \left\{ F(w^k)=\frac{1}{\mathbb{T}_k} \sum_{i\in \mathbb{T}_k} \mbox{\rm KL}(q^i(\bm{\theta})||w^k(\bm{\theta})))\right\}.
	\end{equation}
	Then we have
	\begin{align}
		\mu_{w^k,m}&=\frac{1}{|\mathbb{T}_k|} \sum_{i\in \mathbb{T}_k} \mu_{i,m},\\
		\sigma_{w,m}^2
		&=\frac{1}{|\mathbb{T}_k|} \sum_{i\in \mathbb{T}_k} \left[\sigma_{i,m}^2+\mu_{i,m}^2-\mu_{w^k,m}^2\right].
	\end{align}
\end{lemma}

Due to the definition of $\mathbb{S}_k$, we have
\begin{equation}
	{\hat{k}^i}=k, i \in \mathbb{S}_k.
\end{equation}
So the left-hand term in (50) is equivalent to
\begin{align}
	\frac{1}{|\mathbb{S}_k|} \sum_{i\in \mathbb{S}_k}   \Bigg[  \int_{\Theta} l_n(P^{i},P^{i}_{\bm{\theta}}) \hat{q}^i(\bm{\theta}) d\bm{\theta} 
	+  \zeta\, \mbox{\rm KL}(\hat{q}^i(\bm{\theta})||\hat{w}^k(\bm{\theta})) \Bigg]    
\end{align}

Then by following the techniques from the proof of Lemma 2, it's straightforward to get Lemma 4.

\subsection{Proof of Lemma \ref{lm: optimalsolution}} \label{Proof_Lemma_A2}

The optimality of $w(\bm{\theta})$ in Eq. \eqref{eq: F_w_m} implies that the one-order partial derivatives of $F(w)$ w.r.t. $\mu_{w,m}$ and $\sigma_{w,m}$ are zero
\begin{align} 
	\frac{1}{N} \sum_{i=1}^{N} \frac{\partial \, \mbox{KL}(q^i(\bm{\theta})||w(\bm{\theta}))}{\partial \,\mu_{w,m}} &=0, \label{eq: pd1}\\
	\frac{1}{N} \sum_{i=1}^{N} \frac{\partial \, \mbox{KL}(q^i(\bm{\theta})||w(\bm{\theta}))}{\partial \,\sigma_{w,m}} &=0. \label{eq: pd2}
\end{align}
Using Eq. (9) yields
\begin{align}
	\frac{1}{N} \sum_{i=1}^{N} \frac{\partial \, \mbox{KL}(q^i(\bm{\theta})||w(\bm{\theta}))}{\partial \,\mu_{w,m}} = \frac{1}{N} \sum_{i=1}^{N} \frac{-2(\mu_{i,m}-\mu_{w,m})}{\sigma_{w,m}^2}, \label{eq: pd1a}
\end{align}
and
\begin{multline}
	\frac{1}{N} \sum_{i=1}^{N} \frac{\partial \, \mbox{KL}(q^i(\bm{\theta})||w(\bm{\theta}))}{\partial \,\sigma_{w,m}} \\= \frac{1}{N} \sum_{i=1}^{N} \left[ \frac{2}{\sigma_{w,m}}- \frac{2 \left[\sigma_{i,m}^2+\left(\mu_{i,m}-\mu_{w,m}\right)^2\right]}{\sigma_{w,m}^3} \right].\label{eq: pd2a}
\end{multline}
Therefore, we obtain
\begin{align}
	\mu_{w,m}&=\frac{1}{N} \sum_{i=1}^{N} \mu_{i,m},\\
	\sigma_{w,m}^2&=\frac{1}{N} \sum_{i=1}^{N} \left[\sigma_{i,m}^2+\left(\mu_{i,m}-\mu_{w,m}\right)^2 \right].
\end{align}

\begin{table}[!h]
	\caption{Results of pFedBayes on Medium dataset (MNIST).}
	\label{detail-result-dnn}
	\centering
	\scalebox{1}{
		\begin{tabular}{cccccc}
			\toprule
			\multirow{1}{*}{$\rho$} & $\zeta$ & $\eta_1$ & $\eta_2$ & PM Acc.($\%$)& GM Acc.($\%$)\\
			\toprule
			-1 & 10 & 0.001 & 0.001 & 97.38 & 91.61\\
			-1 & 10 & 0.001 & 0.005 & 97.12 & 90.32\\
			-1 & 10 & 0.005 & 0.001 & 96.73 & 91.34\\
			-1 & 10 & 0.005 & 0.005 & 96.78 & 91.10\\
			\midrule
			-2 & 10 & 0.001 & 0.001 & 97.45 & 92.40\\
			-2 & 10 & 0.001 & 0.005 & 97.42 & 91.35\\
			-2 & 10 & 0.005 & 0.001 & 97.36 & 91.27\\
			-2 & 10 & 0.005 & 0.005 & 97.28 & 90.01\\
			\midrule
			-3 & 10 & 0.001 & 0.001 & 97.08 & 92.16\\
			-3 & 10 & 0.001 & 0.005 & 96.35 & 90.47\\
			-3 & 10 & 0.005 & 0.001 & 96.64 & 87.34\\
			-3 & 10 & 0.005 & 0.005 & 97.28 & 90.22\\
			\midrule
			-1.5 & 10 & 0.001 & 0.001 & 97.38 & 92.31\\
			-2.0 & 10 & 0.001 & 0.001 & 97.45 & 92.40\\
			-2.5 & 10 & 0.001 & 0.001 & 97.18 & 93.22\\
			\midrule
			-2.5 & 0.5 & 0.001 & 0.001 & 97.13 & 89.88\\
			-2.5 & 1 & 0.001 & 0.001 & 97.41 & 91.24\\
			-2.5 & 5 & 0.001 & 0.001 & 97.38 & 93.03\\
			\textbf{-2.5} & \textbf{10} & \textbf{0.001} & \textbf{0.001} & \textbf{97.18} & \textbf{93.22}\\
			-2.5 & 20 & 0.001 & 0.001 & 97.04 & 92.77\\
			\bottomrule
	\end{tabular}}
\end{table}

\begin{table}[ht]
	\centering
	\caption{Results of sFedBayes on Medium dataset (MNIST).}.
	\label{detail-result-dnn-sFedBayes}
	\small
	\scalebox{0.9}{
		\begin{tabular}{ccccccc}
			\toprule
			\textbf{$\rho$} & \textbf{$\zeta$} & \textbf{$\lambda_{\text{init}}$} & \textbf{PM Acc.} & \textbf{PM NNR} & \textbf{GM Acc.} & \textbf{GM NNR} \\
			\midrule
			-2.5 & 10 & 0.1 & 97.21 & 9.02 & 90.16 & 9.07 \\
			-2.5 & 10 & 0.3 & 97.54 & 18.97 & 91.12 & 17.97 \\
			-2.5 & 10 & 0.5 & \textbf{97.60} & 35.58 & \textbf{92.17} & 35.40 \\
			-2.5 & 10 & 0.7 & 96.93 & 66.89 & 91.75 & 66.66 \\
			-2.5 & 10 & 0.9 & 96.49 & 89.94 & 91.10 & 89.93 \\
			\bottomrule
	\end{tabular}}
\end{table}

\section{Hyperparameters Tuning on MNIST Dataset}

We validate pFedBayes and sFedBayes in a basic DNN model with 3 full connection layers [784, 100, 10] on the MNIST Medium dataset, the results are listed in Table~\ref{detail-result-dnn} and Table~\ref{detail-result-dnn-sFedBayes}, respectively.
\textbf{Effects of $\eta_1$ and $\eta_2$:} 
In the pFedBayes algorithm, $\eta_1$ and $\eta_2$ represent the learning rates of the personalized model and the global model, respectively. We tune the learning rate in the range $[0.001, 0.005]$ while fixing other parameters. Table \ref{detail-result-dnn} shows that $\eta_1 = \eta_2 = 0.001$ is the best learning rate setting.
\textbf{Effects of $\zeta$:} 
In the pFedBayes algorithm, $\zeta$ can adjust the degree of personalization.
Therefore, increasing $\zeta$ can improve the performance of the global model, but weaken the performance of the personalized model.
Based on the optimal learning rate, we tune $\zeta \in \{0.5, 1, 5, 10, 20\}$.
From Table~\ref{detail-result-dnn} we can see that $\zeta=10$ is the best setting. 
As a result, $\zeta = 10$ is used for the remaining experiments.

\textbf{Effects of $\rho$:} 
Note that the initialization of weight parameters can affect the results of the model. Therefore, we also tune $\rho \in \{-1, -1.5, -2, -2.5, -3\}$. From Table \ref{detail-result-dnn} we can see that $\rho=-2.5$ is the optimal setting. We set $\rho=-2.5$ for the remaining experiments.
\textbf{Effects of $\lambda$:} 
The value of $\lambda$ represents the initialized sparsity of the model. By controlling $\lambda$, the sparsity of the model can be controlled. However, as the training progresses, the sparsity of the model may change to a certain extent. As can be seen from Table \ref{detail-result-dnn-sFedBayes}, a certain degree of sparsity can speed up model convergence and avoid problems such as overfitting. But when the model is too sparse, the global performance will regress.


\begin{thebibliography}{10}
	\providecommand{\url}[1]{#1}
	\csname url@samestyle\endcsname
	\providecommand{\newblock}{\relax}
	\providecommand{\bibinfo}[2]{#2}
	\providecommand{\BIBentrySTDinterwordspacing}{\spaceskip=0pt\relax}
	\providecommand{\BIBentryALTinterwordstretchfactor}{4}
	\providecommand{\BIBentryALTinterwordspacing}{\spaceskip=\fontdimen2\font plus
		\BIBentryALTinterwordstretchfactor\fontdimen3\font minus
		\fontdimen4\font\relax}
	\providecommand{\BIBforeignlanguage}[2]{{%
			\expandafter\ifx\csname l@#1\endcsname\relax
			\typeout{** WARNING: IEEEtran.bst: No hyphenation pattern has been}%
			\typeout{** loaded for the language `#1'. Using the pattern for}%
			\typeout{** the default language instead.}%
			\else
			\language=\csname l@#1\endcsname
			\fi
			#2}}
	\providecommand{\BIBdecl}{\relax}
	\BIBdecl
	
	\bibitem{mcmahan2017communication}
	B.~McMahan, E.~Moore, D.~Ramage, S.~Hampson, and B.~A. y~Arcas,
	``Communication-efficient learning of deep networks from decentralized
	data,'' in \emph{Artificial Intelligence and Statistics}.\hskip 1em plus
	0.5em minus 0.4em\relax PMLR, 2017, pp. 1273--1282.
	
	\bibitem{Li2020FL}
	T.~Li, A.~K. Sahu, A.~Talwalkar, and V.~Smith, ``Federated learning:
	Challenges, methods, and future directions,'' \emph{IEEE Signal Processing
		Magazine}, vol.~37, no.~3, pp. 50--60, 2020.
	
	\bibitem{li2018federated}
	T.~Li, A.~K. Sahu, M.~Zaheer, M.~Sanjabi, A.~Talwalkar, and V.~Smith,
	``Federated optimization in heterogeneous networks,''
	\emph{Proceedings of Machine Learning and Systems}, vol.~2, pp. 429--450,
	2020.
	
	\bibitem{t2020personalized}
	C.~T~Dinh, N.~Tran, and T.~D. Nguyen, ``Personalized federated learning with
	moreau envelopes,'' \emph{Advances in Neural Information Processing Systems},
	vol.~33, 2020.
	
	\bibitem{hanzely2020federated}
	F.~Hanzely and P.~Richt{\'a}rik, ``Federated learning of a mixture of global
	and local models,'' \emph{arXiv preprint arXiv:2002.05516}, 2020.
	
	\bibitem{li2021personalized}
	Y.~Li, X.~Liu, X.~Zhang, Y.~Shao, Q.~Wang, and Y.~Geng, ``Personalized
	federated learning via maximizing correlation with sparse and hierarchical
	extensions,'' \emph{arXiv preprint arXiv:2107.05330}, 2021.
	
	\bibitem{ghosh2020efficient}
	A.~Ghosh, J.~Chung, D.~Yin, and K.~Ramchandran, ``An efficient framework for
	clustered federated learning,'' \emph{Advances in Neural Information
		Processing Systems}, vol.~33, pp. 19\,586--19\,597, 2020.
	
	\bibitem{xie2020multi}
	G.~Long, M.~Xie, T.~Shen, T.~Zhou, X.~Wang, and J.~Jiang,
	``Multi-center federated learning: Clients clustering for better
	personalization,'' \emph{World Wide Web}, vol.~26, no.~1, pp. 481--500,
	2023.
	
	\bibitem{huang2021personalized}
	Y.~Huang, L.~Chu, Z.~Zhou, L.~Wang, J.~Liu, J.~Pei, and Y.~Zhang,
	``Personalized cross-silo federated learning on non-iid data,'' in
	\emph{Proceedings of the AAAI Conference on Artificial Intelligence},
	vol.~35, no.~9, 2021, pp. 7865--7873.
	
	\bibitem{chen2020fedbe}
	H.-Y. Chen and W.-L. Chao, ``{FedBE: Making Bayesian} model ensemble applicable
	to federated learning,'' in \emph{International Conference on Learning
		Representations}, 2021.
	
	\bibitem{thorgeirsson2021probabilistic}
	A.~T. Thorgeirsson and F.~Gauterin, ``Probabilistic predictions with federated
	learning,'' \emph{Entropy}, vol.~23, no.~1, p.~41, 2021.
	
	\bibitem{jospin2020hands}
	L.~V. Jospin, W.~Buntine, F.~Boussaid, H.~Laga, and M.~Bennamoun, ``Hands-on
	bayesian neural networks-a tutorial for deep learning users,'' \emph{ACM
		Comput. Surv}, vol.~17, no.~2, 2022.
	
	\bibitem{zhang2022personalized}
	X.~Zhang, Y.~Li, W.~Li, K.~Guo, and Y.~Shao, ``Personalized federated learning
	via variational bayesian inference,'' in \emph{International Conference on
		Machine Learning}.\hskip 1em plus 0.5em minus 0.4em\relax PMLR, 2022, pp.
	26\,293--26\,310.
	
	\bibitem{stich2018local}
	S.~U. Stich, ``Local {SGD} converges fast and communicates little,'' in
	\emph{International Conference on Learning Representations}, 2019.
	
	\bibitem{li2019feddane}
	T.~Li, A.~K. Sahu, M.~Zaheer, M.~Sanjabi, A.~Talwalkar, and V.~Smith,
	``{Feddane: A} federated newton-type method,'' in \emph{2019 53rd Asilomar
		Conference on Signals, Systems, and Computers}.\hskip 1em plus 0.5em minus
	0.4em\relax IEEE, 2019, pp. 1227--1231.
	
	\bibitem{zhang2020fedpd}
	X.~Zhang, M.~Hong, S.~Dhople, W.~Yin, and Y.~Liu, ``{Fedpd: A} federated
	learning framework with optimal rates and adaptivity to non-iid data,''
	\emph{arXiv preprint arXiv:2005.11418}, 2020.
	
	\bibitem{guha2019one}
	N.~Guha, A.~Talwalkar, and V.~Smith, ``One-shot federated learning,''
	\emph{arXiv preprint arXiv:1902.11175}, 2019.
	
	\bibitem{sattler2019robust}
	F.~Sattler, S.~Wiedemann, K.-R. M{\"u}ller, and W.~Samek, ``Robust and
	communication-efficient federated learning from non-iid data,'' \emph{IEEE
		transactions on neural networks and learning systems}, vol.~31, no.~9, pp.
	3400--3413, 2019.
	
	\bibitem{rothchild2020fetchsgd}
	D.~Rothchild, A.~Panda, E.~Ullah, N.~Ivkin, I.~Stoica, V.~Braverman,
	J.~Gonzalez, and R.~Arora, ``{Fetchsgd: C}ommunication-efficient federated
	learning with sketching,'' in \emph{International Conference on Machine
		Learning}.\hskip 1em plus 0.5em minus 0.4em\relax PMLR, 2020, pp. 8253--8265.
	
	\bibitem{dai2019hyper}
	X.~Dai, X.~Yan, K.~Zhou, H.~Yang, K.~K. Ng, J.~Cheng, and Y.~Fan,
	``Hyper-sphere quantization: Communication-efficient sgd for federated
	learning,'' \emph{arXiv preprint arXiv:1911.04655}, 2019.
	
	\bibitem{reisizadeh2020fedpaq}
	A.~Reisizadeh, A.~Mokhtari, H.~Hassani, A.~Jadbabaie, and R.~Pedarsani,
	``Fedpaq: A communication-efficient federated learning method with periodic
	averaging and quantization,'' in \emph{International Conference on Artificial
		Intelligence and Statistics}.\hskip 1em plus 0.5em minus 0.4em\relax PMLR,
	2020, pp. 2021--2031.
	
	\bibitem{arivazhagan2019federated}
	M.~G. Arivazhagan, V.~Aggarwal, A.~K. Singh, and S.~Choudhary, ``Federated
	learning with personalization layers,'' \emph{arXiv preprint
		arXiv:1912.00818}, 2019.
	
	\bibitem{smith2017federated}
	V.~Smith, C.-K. Chiang, M.~Sanjabi, and A.~S. Talwalkar, ``Federated multi-task
	learning,'' in \emph{Advances in neural information processing systems},
	2017, pp. 4424--4434.
	
	\bibitem{sattler2021clustered}
	F.~Sattler, K.-R. Müller, and W.~Samek, ``Clustered federated learning:
	Model-agnostic distributed multitask optimization under privacy
	constraints,'' \emph{IEEE Transactions on Neural Networks and Learning
		Systems}, vol.~32, no.~8, pp. 3710--3722, 2021.
	
	\bibitem{chen2018federated}
	F.~Chen, M.~Luo, Z.~Dong, Z.~Li, and X.~He, ``Federated meta-learning with fast
	convergence and efficient communication,'' \emph{arXiv preprint
		arXiv:1802.07876}, 2018.
	
	\bibitem{fallah2020personalized}
	A.~Fallah, A.~Mokhtari, and A.~Ozdaglar, ``Personalized federated learning with
	theoretical guarantees: A model-agnostic meta-learning approach,''
	\emph{Advances in Neural Information Processing Systems}, vol.~33, pp.
	3557--3568, 2020.
	
	\bibitem{xing2022big}
	P.~Xing, S.~Lu, L.~Wu, and H.~Yu, ``Big-fed: Bilevel optimization enhanced
	graph-aided federated learning,'' \emph{IEEE Transactions on Big Data}, 2022.
	
	\bibitem{ma2022convergence}
	J.~Ma, G.~Long, T.~Zhou, J.~Jiang, and C.~Zhang, ``On the convergence of
	clustered federated learning,'' \emph{arXiv preprint arXiv:2202.06187}, 2022.
	
	\bibitem{yurochkin2019bayesian}
	M.~Yurochkin, M.~Agarwal, S.~Ghosh, K.~Greenewald, N.~Hoang, and Y.~Khazaeni,
	``Bayesian nonparametric federated learning of neural networks,'' in
	\emph{International Conference on Machine Learning}.\hskip 1em plus 0.5em
	minus 0.4em\relax PMLR, 2019, pp. 7252--7261.
	
	\bibitem{al2021federated}
	M.~Al-Shedivat, J.~Gillenwater, E.~Xing, and A.~Rostamizadeh, ``Federated
	learning via posterior averaging: A new perspective and practical
	algorithms,'' in \emph{International Conference on Learning Representations},
	2021.
	
	\bibitem{achituve2021personalized}
	I.~Achituve, A.~Shamsian, A.~Navon, G.~Chechik, and E.~Fetaya, ``Personalized
	federated learning with gaussian processes,'' in \emph{Thirty-Fifth
		Conference on Neural Information Processing Systems}, 2021.
	
	\bibitem{cybenko1989approximation}
	G.~Cybenko, ``Approximation by superpositions of a sigmoidal function,''
	\emph{Mathematics of control, signals and systems}, vol.~2, no.~4, pp.
	303--314, 1989.
	
	\bibitem{VIJordan1999}
	M.~I. Jordan, Z.~Ghahramani, T.~S. Jaakkola, and L.~K. Saul, ``An introduction
	to variational methods for graphical models,'' \emph{Machine Learning},
	vol.~37, pp. 183--233, 1999.
	
	\bibitem{Blei2017ReviewVB}
	D.~M. Blei, A.~Kucukelbir, and J.~D. McAuliffe, ``Variational inference: A
	review for statisticians,'' \emph{Journal of the American Statistical
		Association}, vol. 112, no. 518, pp. 859--877, 2017.
	
	\bibitem{Blundell2015}
	C.~Blundell, J.~Cornebise, K.~Kavukcuoglu, and D.~Wierstra, ``Weight
	uncertainty in neural networks,'' in \emph{Proceedings of the 32Nd
		International Conference on International Conference on Machine Learning -
		Volume 37}, ser. ICML'15.\hskip 1em plus 0.5em minus 0.4em\relax JMLR.org,
	2015, pp. 1613--1622.
	
	\bibitem{polson2018posterior}
	N.~G. Polson and V.~Ro{\v{c}}kov{\'a}, ``Posterior concentration for sparse
	deep learning,'' in \emph{Proceedings of the 32nd International Conference on
		Neural Information Processing Systems}, 2018, pp. 938--949.
	
	\bibitem{bai2020efficient}
	J.~Bai, Q.~Song, and G.~Cheng, ``Efficient variational inference for sparse
	deep learning with theoretical guarantee,'' \emph{Advances in Neural
		Information Processing Systems}, vol.~33, 2020.
	
	\bibitem{nakada2020}
	R.~Nakada and M.~Imaizumi, ``Adaptive approximation and generalization of deep
	neural network with intrinsic dimensionality,'' \emph{Journal of Machine
		Learning Research}, vol.~21, no. 174, pp. 1--38, 2020.
	
	\bibitem{cherief2020convergence}
	B.-E. Ch{\'e}rief-Abdellatif, ``Convergence rates of variational inference in
	sparse deep learning,'' in \emph{International Conference on Machine
		Learning}.\hskip 1em plus 0.5em minus 0.4em\relax PMLR, 2020, pp. 1831--1842.
	
	\bibitem{schmidt2020nonparametric}
	J.~Schmidt-Hieber, ``Nonparametric regression using deep neural networks with
	relu activation function,'' \emph{The Annals of Statistics}, vol.~48, no.~4,
	pp. 1875--1897, 2020.
	
	\bibitem{ng2001spectral}
	A.~Ng, M.~Jordan, and Y.~Weiss, ``On spectral clustering: Analysis and an
	algorithm,'' \emph{Advances in neural information processing systems},
	vol.~14, 2001.
	
	\bibitem{von2007tutorial}
	U.~Von~Luxburg, ``A tutorial on spectral clustering,'' \emph{Statistics and
		computing}, vol.~17, no.~4, pp. 395--416, 2007.
	
	\bibitem{lecun2010mnist}
	Y.~LeCun, C.~Cortes, and C.~Burges, ``Mnist handwritten digit database,''
	\emph{ATT Labs [Online]. Available: http://yann.lecun.com/exdb/mnist},
	vol.~2, 2010.
	
	\bibitem{lecun1998gradient}
	Y.~LeCun, L.~Bottou, Y.~Bengio, and P.~Haffner, ``Gradient-based learning
	applied to document recognition,'' \emph{Proceedings of the IEEE}, vol.~86,
	no.~11, pp. 2278--2324, 1998.
	
	\bibitem{xiao2017fashion}
	H.~Xiao, K.~Rasul, and R.~Vollgraf, ``Fashion-mnist: a novel image dataset for
	benchmarking machine learning algorithms,'' \emph{arXiv preprint
		arXiv:1708.07747}, 2017.
	
	\bibitem{krizhevsky2009learning}
	A.~Krizhevsky, ``Learning multiple layers of features from tiny images,''
	\emph{Master's thesis, University of Toronto}, 2009.
	
	\bibitem{netzer2011svhn}
	Y.~Netzer, T.~Wang, A.~Coates, A.~Bissacco, B.~Wu, and A.~Y. Ng, ``Reading
	digits in natural images with unsupervised feature learning,'' \emph{NIPS
		Workshop on Deep Learning and Unsupervised Feature Learning}, 2011.
	
	\bibitem{simonyan2014very}
	K.~Simonyan and A.~Zisserman, ``Very deep convolutional networks for
	large-scale image recognition,'' \emph{arXiv preprint arXiv:1409.1556}, 2014.
	
	\bibitem{NEURIPS2019_bdbca288}
	A.~Paszke, S.~Gross, F.~Massa, A.~Lerer, J.~Bradbury, G.~Chanan, T.~Killeen,
	Z.~Lin, N.~Gimelshein, L.~Antiga, A.~Desmaison, A.~Kopf, E.~Yang, Z.~DeVito,
	M.~Raison, A.~Tejani, S.~Chilamkurthy, B.~Steiner, L.~Fang, J.~Bai, and
	S.~Chintala, ``Pytorch: An imperative style, high-performance deep learning
	library,'' in \emph{Advances in Neural Information Processing Systems},
	H.~Wallach, H.~Larochelle, A.~Beygelzimer, F.~d\textquotesingle
	Alch\'{e}-Buc, E.~Fox, and R.~Garnett, Eds., vol.~32.\hskip 1em plus 0.5em
	minus 0.4em\relax Curran Associates, Inc., 2019.
	
	\bibitem{Osawa2019Practical}
	K.~Osawa, S.~Swaroop, M.~E.~E. Khan, A.~Jain, R.~Eschenhagen, R.~E. Turner, and
	R.~Yokota, ``Practical deep learning with bayesian principles,''
	\emph{Advances in neural information processing systems}, vol.~32, 2019.
	
	\bibitem{Boucheron2013Concentration}
	S.~Boucheron, G.~Lugosi, and P.~Massart, \emph{Concentration inequalities: A
		nonasymptotic theory of independence}.\hskip 1em plus 0.5em minus 0.4em\relax
	Oxford university press, 2013.
	
	\bibitem{pati2018statistical}
	D.~Pati, A.~Bhattacharya, and Y.~Yang, ``On statistical optimality of
	variational bayes,'' in \emph{International Conference on Artificial
		Intelligence and Statistics}.\hskip 1em plus 0.5em minus 0.4em\relax PMLR,
	2018, pp. 1579--1588.
	
\end{thebibliography}
\end{document}